\documentclass[runningheads]{llncs}

\PassOptionsToPackage{table}{xcolor}

\usepackage{eccv}

\usepackage{eccvabbrv}

\usepackage{graphicx}
\usepackage{booktabs}
\usepackage{tabularx}
\usepackage{multirow}
\usepackage{amsfonts}       
\usepackage{enumitem}       

\usepackage[accsupp]{axessibility}

\DeclareMathOperator{\E}{\mathbb{E}}
\DeclareMathOperator{\SSIM}{SSIM}

\usepackage[breaklinks,colorlinks,citecolor=blue,linkcolor=black,urlcolor=blue]{hyperref}

\begin{document}

\title{AsyncEvGS: Asynchronous Event-Assisted Gaussian Splatting for Handheld Motion-Blurred Scenes}

\titlerunning{AsyncEvGS}

\author{Jun Dai\inst{1}\textsuperscript{*} \and  
Renbiao Jin\inst{2}\textsuperscript{*} \and          
Bo Xu\inst{2} \and
Yutian Chen\inst{3} \and
Linning Xu\inst{3} \and
Mulin Yu\inst{1} \and
Tianfan Xue\inst{3,1,4}\textsuperscript{\dag} \and   
Shi Guo\inst{1}\textsuperscript{\dag}}           

\authorrunning{J.~Dai et al.}

\institute{Shanghai AI Laboratory \and
Shanghai Jiao Tong University \and
CUHK MMLab \and
CPII under InnoHK\\[2pt]
\textsuperscript{*}Equal contribution.\quad
\textsuperscript{\dag}Corresponding author.\\
\email{jundai332@gmail.com, guoshi@pjlab.org.cn}}

\maketitle

\begin{figure}[t]
    \centering
    \includegraphics[width=\textwidth]{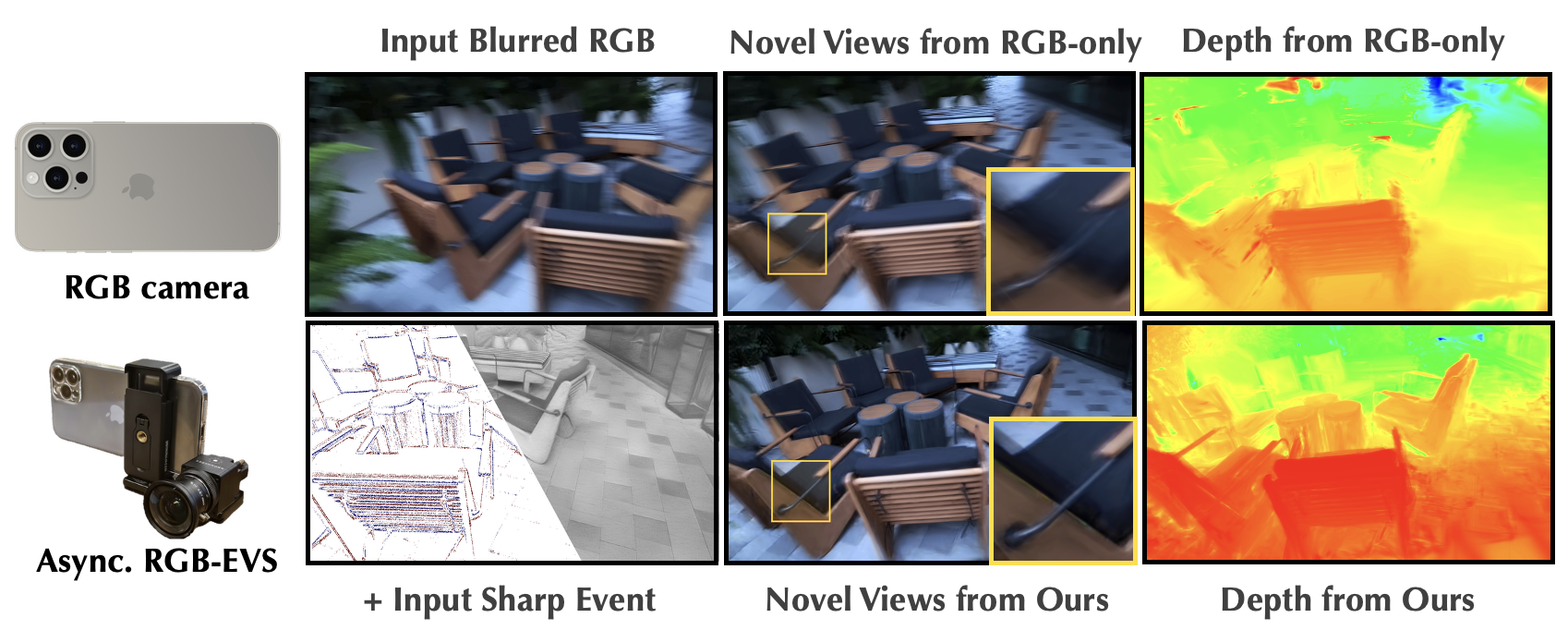}
    \caption{
    High-quality 3D reconstruction from severely blurred inputs captured during rapid handheld motion. (Top) Reconstructing from blurred RGB images alone is ill-posed: RGB-only methods (BAGS~\cite{peng2024bags}) fail to resolve motion ambiguity, producing blurry novel views with noticeable artifacts and distorted geometry.
    (Bottom) We propose a high-resolution asynchronous RGB--EVS system that pairs a handheld RGB camera with an event sensor. Leveraging the sharp, high-frequency cues from events, our method recovers accurate geometry and produces sharp novel views. This shows that \textit{ our system can effectively use event signals to boost 3D reconstruction quality for commonly used handheld RGB cameras such as smartphones.}
    }
    \label{fig:teaser}
\end{figure}

\begin{abstract}
3D reconstruction methods such as 3D Gaussian Splatting (3DGS) and Neural Radiance Fields (NeRF) achieve impressive photorealism but fail when input images suffer from severe motion blur. While event cameras provide high-temporal-resolution motion cues, existing event-assisted approaches rely on low-resolution sensors and strict synchronization, limiting their practicality for handheld 3D capture on common devices, such as smartphones. We introduce a flexible, high-resolution \textbf{asynchronous} RGB–Event dual-camera system and a corresponding reconstruction framework. Our approach first reconstructs sharp images from the event data and then employs a cross-domain pose estimation module based on the Visual Geometry Transformer (VGGT) to obtain robust initialization for 3DGS. During optimization, we employ a structure-driven event loss and view-specific consistency regularizers to mitigate the ill-posed behavior of traditional event losses and deblurring losses, ensuring both stable and high-fidelity reconstruction. We further contribute AsyncEv-Deblur, a new high-resolution RGB–Event dataset captured with our asynchronous system. Experiments demonstrate that our method achieves state-of-the-art performance on both our challenging dataset and existing benchmarks, substantially improving reconstruction robustness under severe motion blur. Project page: \url{https://openimaginglab.github.io/AsyncEvGS/}.
\end{abstract}

\section{Introduction}
\label{sec:intro}

Neural Radiance Fields (NeRF)~\cite{mildenhall2021nerf} and 3D Gaussian Splatting (3DGS)~\cite{kerbl20233d} have recently achieved unprecedented photorealism in novel view synthesis. Their success, however, hinges on a collection of high-quality, sharp input images---an assumption frequently violated in real-world 3D scanning. Handheld capture, in particular, is often plagued by severe motion blur~\cite{lee2023dp, zhao2024bad, ma2022deblur}, as also shown in Fig~\ref{fig:teaser}. Given the prevalence of handheld devices in 3D capture and robotics, it is critical to enhance reconstruction robustness under motion blur.
While computational deblurring 3DGS techniques~\cite{wang2023bad,zhao2024bad,ma2022deblur,peng2024bags} exist, they grapple with the ill-posed nature of deblurring. Event cameras, in contrast, offer a powerful alternative~\cite{bauersfeld2025monoculareventcameramotioncapture, Gehrig24nature}. With their high temporal resolution and asynchronous measurement of intensity changes, they provide robust motion cues even in the presence of severe blur~\cite{qi2023e2nerf, tang2025lse, Matta_2025_WACV, deguchi2024e2gseventenhancedgaussian, huang2025inceventgs, rudnev2023eventnerf}.

However, existing event-assisted 3D reconstruction methods face two critical limitations that preclude their use in common handheld scenarios.
(1) \textbf{Low Sensor Resolution.} Most methods~\cite{rudnev2023eventnerf, qi2023e2nerf, huang2025inceventgs, qi2024deblurring, klenk2023nerf, ma2023deformable, rudnev2025dynamiceventnerf} are built upon low-resolution event sensors (\eg, DAVIS $346\times260$). This resolution, substantially lower than modern multi-megapixel RGB cameras, fundamentally limits the achievable reconstruction fidelity~\cite{wang2021nerf-sr}.
(2) \textbf{Temporal Synchronization Requirement.} While high-resolution stereo setups like LSENeRF~\cite{tang2025lse} address the resolution bottleneck, they mandate rigid, hardware-level temporal synchronization. This reliance on external triggering restricts their use to specialized industrial cameras and is incompatible with prevalent handheld devices (\eg, smartphones, RealSense~\cite{tadic2019application}) where such synchronization is unavailable. 
This disparity raises a critical question: \emph{Can event cameras be leveraged to deblur 3D reconstructions from common, unsynchronized handheld devices?}

To address these challenges, we introduce a simple but novel, high-resolution \textbf{asynchronous} RGB-Event dual-camera rig (Fig.~\ref{fig:teaser}).
Without hardware synchronization, the rig can be readily paired with a commodity RGB camera for in-the-wild capture.
However, asynchrony breaks a key assumption in prior RGB–Event systems~\cite{tang2025lse}: event-camera poses can no longer be obtained by directly transferring RGB poses via a fixed extrinsic calibration.
As a practical solution, we build a two-stage pipeline that couples event-to-intensity reconstruction with cross-modal pose estimation to obtain reliable poses for both RGB and Event.
We first reconstruct intensity images from the event stream using E2VID~\cite{Rebecq19cvpr}. Although E2VID provides sharp grayscale cues (Fig.~\ref{fig:teaser}), COLMAP~\cite{Schonberger_2016_CVPR} remains brittle when jointly registering motion-blurred RGB frames and event reconstructions, often yielding fragmented trajectories.
Motivated by recent progress in feed-forward 3D foundation models, we leverage VGGT~\cite{wang2025vggt} to obtain robust cross-modal pose initialization~\cite{chendeep}; using the E2VID reconstructions as an event-derived bridge, this initialization enables stable 3D Gaussian Splatting reconstruction in our asynchronous setting.

Beyond pose initialization, effective data fusion during optimization is also critical. Prior event-assisted methods~\cite{rudnev2023eventnerf, qi2023e2nerf} often rely on \emph{cross-view} supervision. Without direct per-view constraints, this approach renders the appearance estimation problem inherently ill-posed, especially in our asynchronous setting, where each view contains only a single modality, \textit{either} an RGB image \textit{or} an event-derived grayscale observation.
We address this by introducing an optimization framework that augments standard event losses~\cite{tang2025lse,klenk2023nerf,yu2024evagaussians} with a structure-based loss. Guided by an event-confidence map computed via multi-scale high-frequency consistency, this loss selectively emphasizes reliable event structures while suppressing noisy or uninformative regions, enabling effective exploitation of high-frequency cues from the event. 
Furthermore, existing deblur modules~\cite{ma2022deblur,qi2024deblurring,zhao2024bad} constrain a blurred observation using an averaging of renderings from neighboring views, which can admit degenerate “compensation” solutions: errors in individual views may cancel out after aggregation and still match the blurred image. To prevent this, we introduce a consistency regularizer for RGB views that encourages neighboring latent appearances to be consistent, reducing artifacts and stabilizing optimization under pose noise.



Since there are no asynchronous event-RGB data for 3D reconstruction, we also have collected  \textbf{AsyncEv-Deblur}, a new RGB-EVS dataset that we will release. This dataset features diverse scenes captured with our high-resolution setup. Our experiments demonstrate that our method not only excels on this new, challenging dataset but also achieves state-of-the-art performance on public benchmarks.
In summary, our contributions are threefold:

\begin{itemize}[leftmargin=*,itemsep=2pt,topsep=3pt]
    \item \textbf{A Novel High-Resolution Asynchronous System:} We propose the first practical pipeline for high-fidelity 3D reconstruction using a flexible $1280 \times 720$ RGB-Event rig, overcoming the low-resolution (\eg, $346 \times 260$) limits of prior work.
    \item \textbf{A Robust Cross-Domain Algorithmic Framework:} We introduce a robust initialization pipeline using VGGT for cross-domain pose estimation, and a tailored optimization framework featuring a novel event structure loss and consistency regularizers.
    \item \textbf{A New Benchmark Dataset and SOTA Performance:} We present AsyncEv-Deblur, a new high-resolution RGB-Event dataset, and demonstrate that our method achieves state-of-the-art performance, significantly outperforming existing approaches.
\end{itemize}
\section{Related work}
\subsection{Deblurring 3D reconstruction }
Recent research has actively explored deblurring neural rendering to recover sharp and geometrically consistent 3D scenes from motion-blurred inputs. Conventional neural rendering frameworks such as NeRF~\cite{mildenhall2021nerf} and 3D Gaussian Splatting (3DGS)~\cite{kerbl20233d} assume static scenes and sharp multi-view images, leading to severe reconstruction artifacts when exposed to motion blur. To address this, 
several trajectory-based deblurring pipelines have been introduced.

Trajectory-based approaches explicitly model the motion trajectory of the camera or scene during exposure. BAD-NeRF~\cite{wang2023bad}, ExBluRF~\cite{lee2023exblurf}, DyBluRF~\cite{sun2024dyblurf}, and BAD-Gaussians~\cite{zhao2024bad} jointly optimize the latent 3D representation and exposure trajectory by synthesizing blurred renderings through temporal integration of multiple sharp latent images. Later methods further extend this paradigm within the Gaussian Splatting framework. CRiM-GS~\cite{lee2024crim} and CoMoGaussian~\cite{lee2025comogaussian} adopt continuous-time neural ODEs to parameterize camera trajectories, achieving smoother and more flexible motion modeling. BARD-GS~\cite{lu2025bard}, MoBGS~\cite{bui2025mobgs}, and MoBluRF~\cite{bui2025moblurf} enhance dynamic scene modeling by disentangling static and moving regions, enabling temporally coherent novel view synthesis under severe motion blur.

Other methods jointly optimize geometric and radiance attributes to better handle spatially varying blur. Deblur-NeRF~\cite{ma2022deblur}, PDRF~\cite{peng2022pdrf}, and DP-NeRF~\cite{lee2023dp} incorporate differentiable blur kernels and depth-dependent transformations into the rendering process to simulate the blur formation model. Within explicit Gaussian representations, BAGS~\cite{peng2024bags} and Deblurring 3DGS~\cite{lee2024deblurring} refine per-Gaussian anisotropy to adaptively encode the blur field, while DeepDeblurRF~\cite{choi2025exploiting} integrates pretrained 2D deblurring priors into the 3D radiance field optimization. However, due to the ill-posed nature of the blur formation, these methods struggle to handle large motion blurs effectively.

\subsection{Event-based  deblurring 3D reconstruction}


Event-based sensors offer microsecond-level temporal resolution and high dynamic range, making them ideal for mitigating motion blur and lighting saturation. Recent methods leverage event streams to guide 3D reconstruction from degraded inputs. E-NeRF~\cite{klenk2023nerf} formulates NeRF training using event generation-based supervision, comparing predicted brightness changes with real event streams. E2NeRF~\cite{qi2023e2nerf} and Ev-DeblurNeRF~\cite{cannici2024mitigating} employ the Event-based Double Integral (EDI) model~\cite{pan2019bringing} to reconstruct sharp latent frames for pose initialization and consistent radiance learning. Moving to explicit representations, EvaGaussians~\cite{yu2024evagaussians}, EaDeblur-GS~\cite{weng2024eadeblur}, and DiET-GS~\cite{lee2025diet} integrate event-based temporal priors and EDI-guided supervision into 3DGS optimization, jointly refining motion trajectory, event consistency, and Gaussian attributes for high-fidelity reconstruction.
These advances demonstrate that integrating event signals into neural rendering provides physically grounded constraints that effectively mitigate motion blur, ensuring temporally precise and geometrically stable 3D reconstruction in dynamic real-world environments. 
The capture hardware differences are summarized in~\cref{tab:method_comparison}.
Existing event-assisted systems are limited by low sensor resolution and strict temporal synchronization, restricting their use in practical, high-resolution scenarios such as mobile capture.
To address this, we propose an asynchronous RGB-Event solution for high-quality 3D reconstruction.


        

\begin{table}[t!]
    \centering 
    \caption{Comparison of key hardware configurations. We compare against DeblurGS~\cite{lee2024deblurring}, E2NeRF~\cite{qi2023e2nerf}, and LSENeRF~\cite{tang2025lse}. \textit{Both} denotes the use of RGB and Event cameras. \textit{Temp Sync.} denotes if strict temporal synchronization is required between the event camera and the RGB camera.} 
    
    \label{tab:method_comparison} 
    \newcolumntype{Y}{>{\centering\arraybackslash}X}
    \begin{tabularx}{\linewidth}{m{0.16\linewidth}YYYY}
        \toprule
        &
        {\footnotesize DeblurGS} &
        {\footnotesize E2NeRF} &
        {\footnotesize LSENeRF} &
        {\footnotesize \textbf{Ours}} \\
        \midrule

        \scalebox{0.9}{\emph{Cam Type.}} & \scalebox{0.9}{RGB}  & \scalebox{0.9}{Both}  & \scalebox{0.9}{Both}  & \scalebox{0.9}{Both}  \\

        \scalebox{0.9}{\emph{Resolution}} & \scalebox{0.8}{$600\times400$}  & \scalebox{0.8}{$346\times260$}  & \scalebox{0.8}{$1280\times720$}  & \scalebox{0.8}{$1280\times720$} \\

        \scalebox{0.9}{\emph{Temp Sync.}} & {-} & \scalebox{0.9}{Yes} & \scalebox{0.9}{Yes} & \scalebox{0.9}{No}  \\
        \bottomrule
    \end{tabularx}
\end{table}

\begin{figure}[tb]
  \centering
   \includegraphics[width=\textwidth]{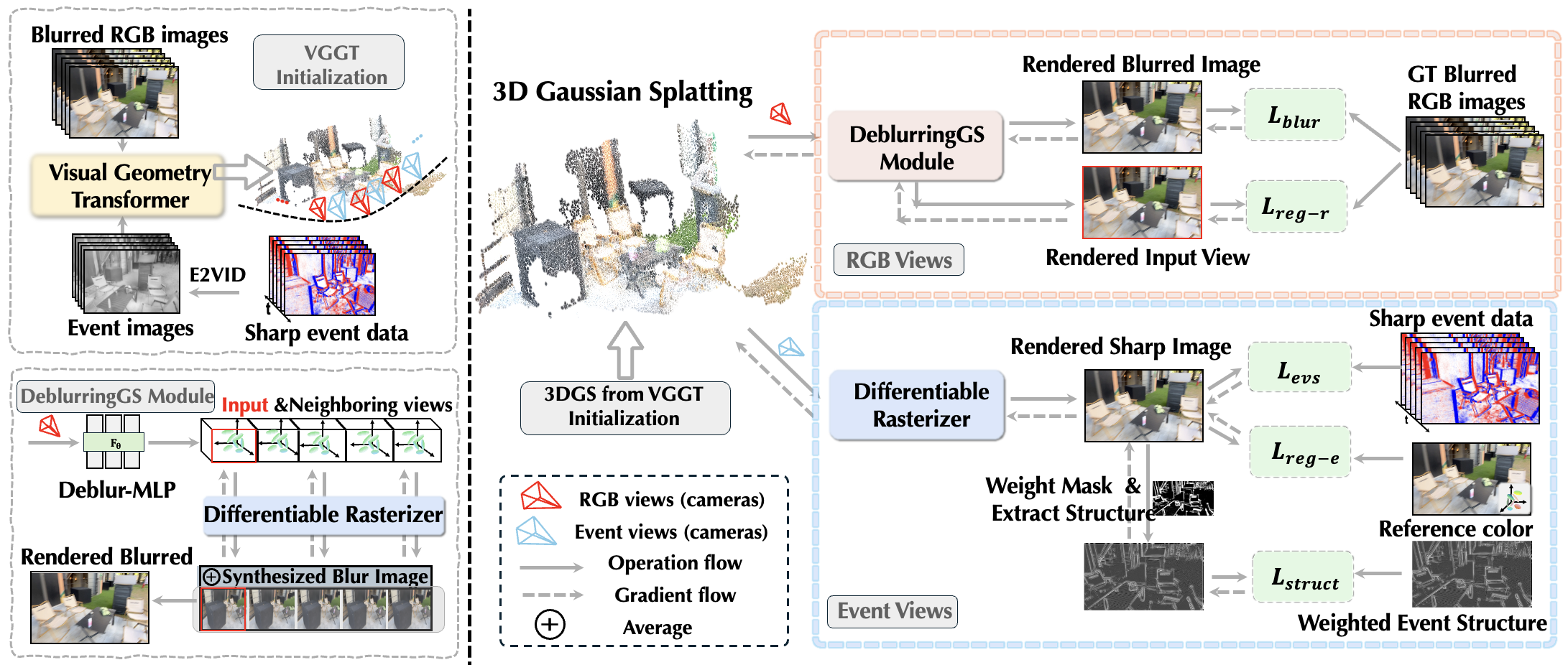}

    \caption{An overview of our proposed reconstruction pipeline. Our method takes blurred RGB images and sharp event streams as input. We first employ VGGT~\cite{wang2025vggt} to process both RGB and event images, providing robust initial camera poses and 3DGS points. The 3DGS representation is then jointly optimized using five key losses, broadly categorized into three groups:
    \textbf{(1) Deblurring Losses:} The blur synthesis loss ($\mathcal{L}_{\text{blur}}$) matches the synthesized blur to the input, while an RGB consistency regularizer ($\mathcal{L}_{\text{reg-r}}$) prevents degradation of the sharp neighboring views.
    \textbf{(2) Event-Guided Losses:} We augment the traditional photometric loss ($\mathcal{L}_{\text{evs}}$), with our novel structure loss ($\mathcal{L}_{\text{struct}}$) to robustly leverage high-frequency event details.
    \textbf{(3) Consistency Loss ($\mathcal{L}_{\text{reg-e}}$):}
    A color distillation loss ensures that event views match the colors learned from a coarse (Stage 1) 3DGS copy.
    }
   \label{fig:pipeline}
\end{figure}
\section{Method}
We propose an asynchronous RGB-Event system for high-fidelity 3D reconstruction from motion-blurred RGB images and sharp event data. Our pipeline, illustrated in ~\cref{fig:pipeline}, is structured around two core stages: 1) a robust cross-domain initialization and pose estimation framework that bypasses traditional SfM tools (e.g., COLMAP), and 2) a specialized optimization framework for 3D Gaussian Splatting. For optimization, we introduce a multi-objective loss function. This includes a deblurring loss for the RGB data, which we augment with a novel \textbf{event structure loss} to enforce high-frequency details. Concurrently, we employ a \textbf{consistency regularization term} to prevent color degradation. We will detail each of these components in the subsequent sections.

\subsection{RGB-Event dual-camera system}
\label{sec:system} 
Our 3D reconstruction pipeline is fed by a high-resolution, dual-camera capture system (shown in Fig.~\ref{fig:teaser}). This design addresses the significant resolution gap between common event sensors (\eg, DAVIS346) and the high-definition images required by NeRF or 3DGS. Our system comprises: (1) a Prophesee EVK-3 HD event camera to capture high-resolution ($1280 \times 720$) event streams, and (2) a separate RGB camera (\ie, an iPhone 13, set to $1280 \times 720$) to provide the information for colorful reconstruction. Critically, our system operates in a much more flexible way; it requires no hardware-synchronization and can use high-resolution RGB sensor (\cref{tab:method_comparison}). This is enabled by our novel initialization method (Sec.~\ref{sec:initialization}), which allows the cameras to be flexibly co-mounted on a simple rigid bracket.

\subsection{Camera poses and 3DGS initialization}
\label{sec:initialization} 
Estimating camera poses for our dual-camera system is non-trivial. A conventional approach might derive event camera poses from the RGB camera's COLMAP estimates via a pre-calibrated relative extrinsic between the RGB and Event cameras~\cite{tang2025lse}. However, this not only requires meticulous pre-calibration of both the relative extrinsics and COLMAP's global scale but also mandates temporal synchronization. An alternative is to convert event streams to gray-scale frames via models like E2VID~\cite{Rebecq19cvpr} for subsequent COLMAP processing. However, the joint calibration struggles due to severe motion blur in the RGB frames and the inherent domain gap between the RGB and reconstructed gray-scale images.

To ensure flexibility and achieve robust, efficient pose estimation, we leverage VGGT~\cite{wang2025vggt} for joint calibration. Benefiting from its strong data priors, VGGT can process challenging, motion-blurred RGB inputs while producing a denser and more accurate initialization for 3D Gaussian Splatting compared to COLMAP. Specifically, we first convert the raw event stream into a sequence of sharp gray-scale images using E2VID. These images are then post-processed with bilateral denoising and multi-frame brightness equalization. Finally, we feed both the blurred RGB frames and the sharp gray-scale frames into VGGT. This process yields a dense point cloud, which serves as the 3DGS initialization, along with the corresponding camera poses for all input images. Compared to COLMAP, our initialization method is significantly more robust to severe motion blur and provides a more accurate initial geometry as shown in \cref{fig:dense_init}.

\begin{figure}[!t]
  \centering
  \begin{subfigure}[t]{0.48\textwidth}
    \centering
    \includegraphics[width=\linewidth]{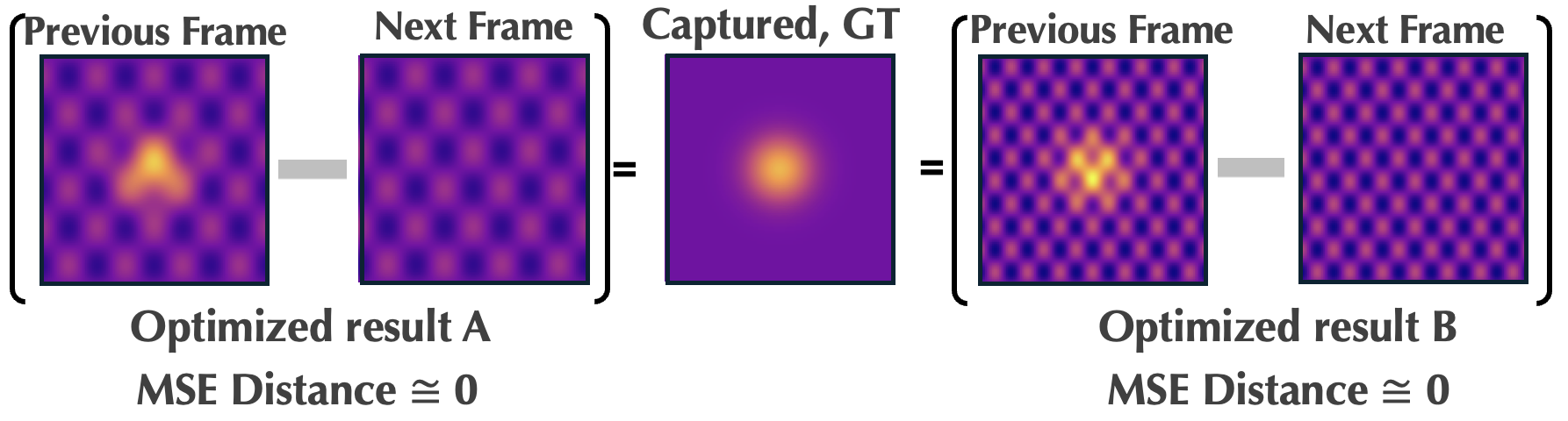}
    \caption{2D toy example of the ill-posed classical event loss. It renders two consecutive frames and minimizes the difference against acquired event data, which is prone to ambiguity and limits reconstruction quality.}
    \label{fig:event-loss-illposed}
  \end{subfigure}
  \hfill
  \begin{subfigure}[t]{0.48\textwidth}
    \centering
    \includegraphics[width=\linewidth]{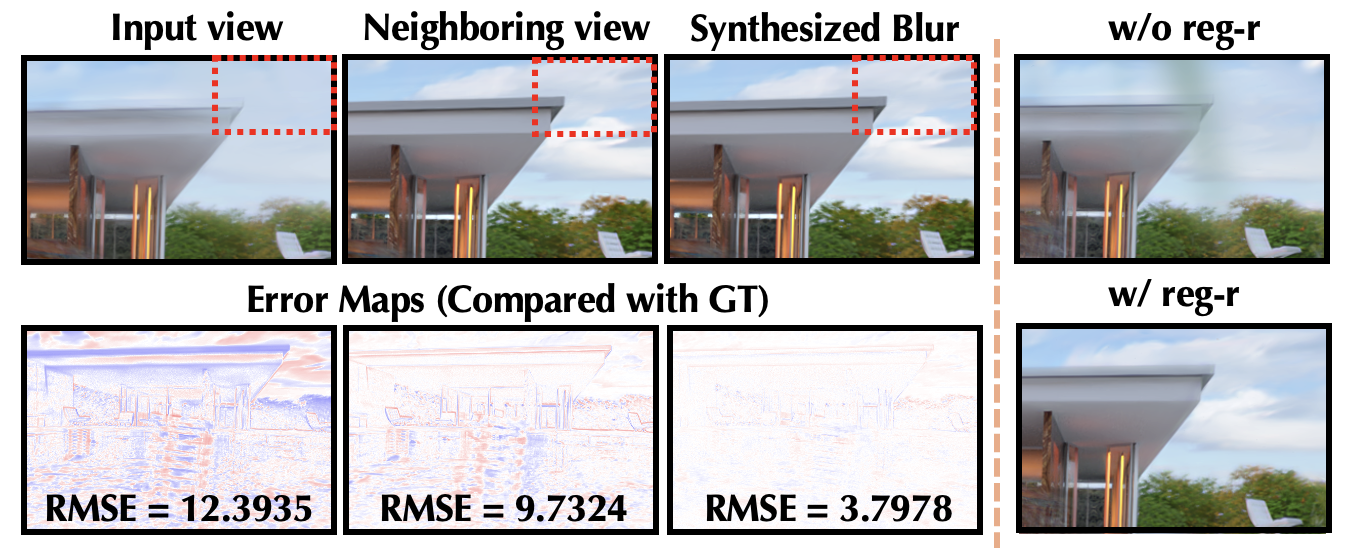}
    \caption{The blur synthesis loss is ill-posed. The model can achieve a low loss by rendering accurate \textit{Neighboring views} while producing artifacts in the (training) \textit{Input view} (``w/o reg.''). Our regularization recovers a sharper result (``w/ reg.'').}
    \label{fig:event-loss-dis}
  \end{subfigure}
  \caption{Illustration of ill-posed problems in our optimization.
    \textbf{(a)} The classical event loss only constrains the intensity \textit{difference} between adjacent frames, providing no absolute supervision and resulting in limited reconstruction quality.
    \textbf{(b)} Synthesizing motion blur by averaging neighboring views can also converge to a degenerate solution; our consistency regularizer effectively mitigates this.}
  \label{fig:illposed-combined}
\end{figure}

\subsection{Event Structure Loss}
\label{sec:event_loss}
Conventional event-based 3D reconstructions~\cite{rudnev2023eventnerf, qi2023e2nerf, tang2025lse} employ event-based losses, such as the photometric consistency loss, to supervise the change of log-intensity between two adjacent timestamps $t_s$ and $t_e$:
\begin{equation}
\label{eq:event_loss_multiline}
\small
\mathcal{L}_{\text{evs}} = \mathbb{E}_{t_s,t_e}
\Big[ \big\|
\log I(t_e) - \log I(t_s) - c\!\sum\nolimits E(t_s,t_e)
\big\|_2^2 \Big],
\end{equation}
where $c$ represents the contrast threshold that triggers events, and $E$ denotes the event signal. This formulation is inherently ill-posed, as it only constrains the \textit{difference} and provides no supervision for the absolute intensity.
Such ill-posedness limits the recovery of fine textures, as illustrated in~\cref{fig:event-loss-illposed}.
While using an event-to-video network (\eg, E2VID~\cite{Rebecq19cvpr}) to generate grayscale “ground truth” images $I_{\text{GT}}$ for direct supervision can alleviate this ambiguity, the E2VID outputs suffer from significant brightness inconsistencies—both across frames and within weakly textured regions.
Naively supervising on these images would bake these artifacts directly into the 3DGS, leading to severe degradation.

To resolve this, we propose an \textbf{Event Structure Loss}, $\mathcal{L}_{\text{struct}}$, designed to be robust to both brightness artifacts and potential pose inaccuracies.
First, motivated by the fact that events are most reliable at edges, we isolate high-frequency structural information using a structure extractor $\mathrm{S}(\cdot)$.
We then derive a confidence map $W$ from the cross-scale consistency of $\mathrm{S}$, where structurally consistent regions receive high confidence and textureless areas are down-weighted.
Second, we must account for small pose inaccuracies from our estimator, which can cause minor view shifts.
Therefore, we compute our loss using the Structural Similarity (SSIM) index~\cite{1284395}, which is inherently robust to small translations and focuses on structural correctness rather than unstable pixel-wise alignment.

Our final event structure loss is defined as a weighted SSIM, computed as the expectation over all pixels $p$:
\begin{equation}
\small
\label{eq:weighted_ssim}
\mathcal{L}_{\text{struct}} = 1 - \E_{p} \Big[ W(p) \cdot \SSIM \big( \mathrm{S}(I_{\text{ren}}), \mathrm{S}(I_{\text{evs}}) \big) (p) \Big],
\end{equation}
where $I_{\text{ren}}$ denotes the Gaussian-rendered image, $I_{\text{evs}}$ is the E2VID-reconstructed target, $\mathrm{S}(\cdot)$ is the structure extractor (following~\cite{Chen_2025_CVPR}), $W(p)$ is the confidence weight at pixel $p$, and $\SSIM(\cdot)(p)$ computes the pixel-wise SSIM value. Note that the luminance term is removed from SSIM to eliminate brightness inconsistencies between RGB and event-based grayscale images.
%


\subsection{Consistency regularization}
\label{sec:color_reg}

To handle motion blur in the RGB inputs, we adopt the deblurring strategy from Deblurring 3DGS~\cite{lee2024deblurring}.
This method uses an MLP to estimate per-view offsets and render $N$ sharp neighboring images ${I}_{i=1}^N$ for each RGB view.
These images are averaged to synthesize a blurred image $I_{\text{avg}} = \frac{1}{N}\sum_{i=1}^N I_i$, which is supervised by comparing with the captured blurred image $I'$,
\begin{equation}
\small
\label{eq:motion_blur_loss}
\mathcal{L}_{\text{blur}} = (1 - \alpha) \cdot \| I_{\text{avg}} - I' \|_1 + \alpha \cdot \big(1 - \SSIM(I_{\text{avg}}, I')\big),
\end{equation}
where $\alpha$ denotes the loss weighting factor.

However, we find this deblurring loss is ill-posed on its own, as it only enforces cross-view constraints.
As shown in \cref{fig:event-loss-dis}, 3DGS may converge to a local optimum, causing the rendered input training views to exhibit severe degradation.
The incorporation of event signals further amplifies this instability.
To mitigate this ill-posedness, we introduce consistency regularizers on both the RGB and event views, denoted as $\mathcal{L}_{\text{reg-r}}$ and $\mathcal{L}_{\text{reg-e}}$, respectively.


For the RGB views, we impose additional constraints to stabilize the rendered image.
We further enforce that the rendered image itself should remain close to both the neighboring sharp frames and the observed blurred input:
\begin{equation}
\label{eq:reg_1}
\small
\mathcal{L}_{\text{reg-r}} =
\frac{1}{N}\sum_{i=1}^{N} \big\| g(I_i) - g(I) \big\|_2^2
+
\big\| g(I) - I' \big\|_2^2,
\end{equation}
where $g(\cdot)$ denotes Gaussian blurring, $I_i$ is the $i$-th neighboring sharp frame estimated by the deblurring MLP module, $I'$ is the blurry observation.

For the event views, we aim to introduce a color constraint to prevent the rendered appearance from drifting.
To obtain such a reference color prior, we first train a coarse Gaussian model $\mathcal{G}_{\text{ref}}$ using only the RGB images.
Although this model provides poor structural quality, it still provides a rough but reliable color reference.
Thus, for an event-view pose $P$, the corresponding regularizer $\mathcal{L}_{\text{reg-e}}$ is defined as:
\begin{equation}
\label{eq:reg_2}
\small
\mathcal{L}_{\text{reg-e}} = \mathbb{E}_{P \in \mathcal{R}_{\text{evt}}}
\Big[ \big\|
I(P) - \mathcal{G}_{\text{ref}}(P)
\big\|_2^2 \Big],
\end{equation}
where $P$ denotes a pose from the set of event-view poses $\mathcal{R}_{\text{evt}}$, and $I(P)$ is the rendered image.

\section{Experiments}

\subsection{Implementation Details}
\paragraph{Training Details.}
We implement our method based on the Gsplat framework~\cite{ye2025gsplat} and adopt the RGB deblurring module from Deblurring 3DGS~\cite{lee2024deblurring}.
In Stage 1, we train a 3DGS using only RGB images, initialized from VGGT-reconstructed 3D points filtered by a 50\% confidence threshold. We render $N=5$ views to synthesize the final blurred image.
The model is optimized for $10k$ iterations using Adam~\cite{kingma2015adam} to minimize the blur loss $\mathcal{L}_{\mathrm{blur}}$.
The learning rates are default setting from Gsplat.
Following this, in Stage 2, the 3DGS from Stage 1 is copied to serve as a fixed reference for color supervision. We then train a new 3DGS ($30k$ iters), again initialized from VGGT points, using both RGB and event data.
This model is optimized using Adam with our multi-objective loss.
We set the loss weights to $\lambda_{\text{blur}} = 1.0$, $\lambda_{\text{struct}} = 0.2$, $\lambda_{\text{evs}} = 0.002$, $\lambda_{\text{reg-r}} = 0.2$, and $\lambda_{\text{reg-e}} = 1.0$.
The learning rates are identical to those in Stage 1.
Further details regarding the structure extractor and the deblurring MLP are provided in the supplementary material.

\paragraph{Evaluation datasets.} We evaluate our method on our newly captured real-world \textbf{AsyncEv-Deblur} dataset and a modified \textbf{Ev-DeblurBlender} dataset~\cite{cannici2024mitigating}. 
Our AsyncEv-Deblur dataset contains 7 scenes: \textit{Patio, Bin, Lounge, Bench, Stair, Bus,} and \textit{Wall}. 
For each scene, we perform a rapid handheld sweep using the RGB camera and the event sensor to capture the blurred inputs at the same time, 
followed by a slow, stable pass with the RGB camera alone to record high-quality ground-truth images.
For the Ev-DeblurBlender dataset, we utilize all four scenes: \textit{factory, pool, tanabata,} and \textit{trolley}. Crucially, as VGGT initialization is vital for reconstruction, we use our VGGT pipeline to re-calibrate all camera poses and 3DGS initializations for both datasets, ensuring an unbiased comparison. Further details are provided in the supplementary material.

\paragraph{Baselines.} We evaluate our method against two categories of baselines: RGB-only and RGB-Event fusion methods. For the RGB-only category, we select the original 3DGS \cite{kerbl20233d}, BAGS \cite{peng2024bags}, and DeblurringGS \cite{lee2024deblurring} as recent state-of-the-art deblurring and reconstruction methods. For the RGB-Event category, finding a directly comparable baseline is challenging due to our use of an asynchronous dual-camera system. Most existing work employs low-resolution, single-camera setups (\eg, DAVIS346). To the best of our knowledge, LSENeRF \cite{tang2025lse} is the only other work utilizing a dual-camera system. However, its requirement for strict camera synchronization, which our setup lacks, results in incompatible data formats. Therefore, we adapt its event loss configuration to our data format and re-implement it in our codebase. To ensure a fair comparison, all baselines uniformly utilize our camera poses and 3DGS initialization calibrated by VGGT.

\subsection{Experimental Validations}
\begin{figure}[tb]
  \centering
   \includegraphics[width=1.0\linewidth]{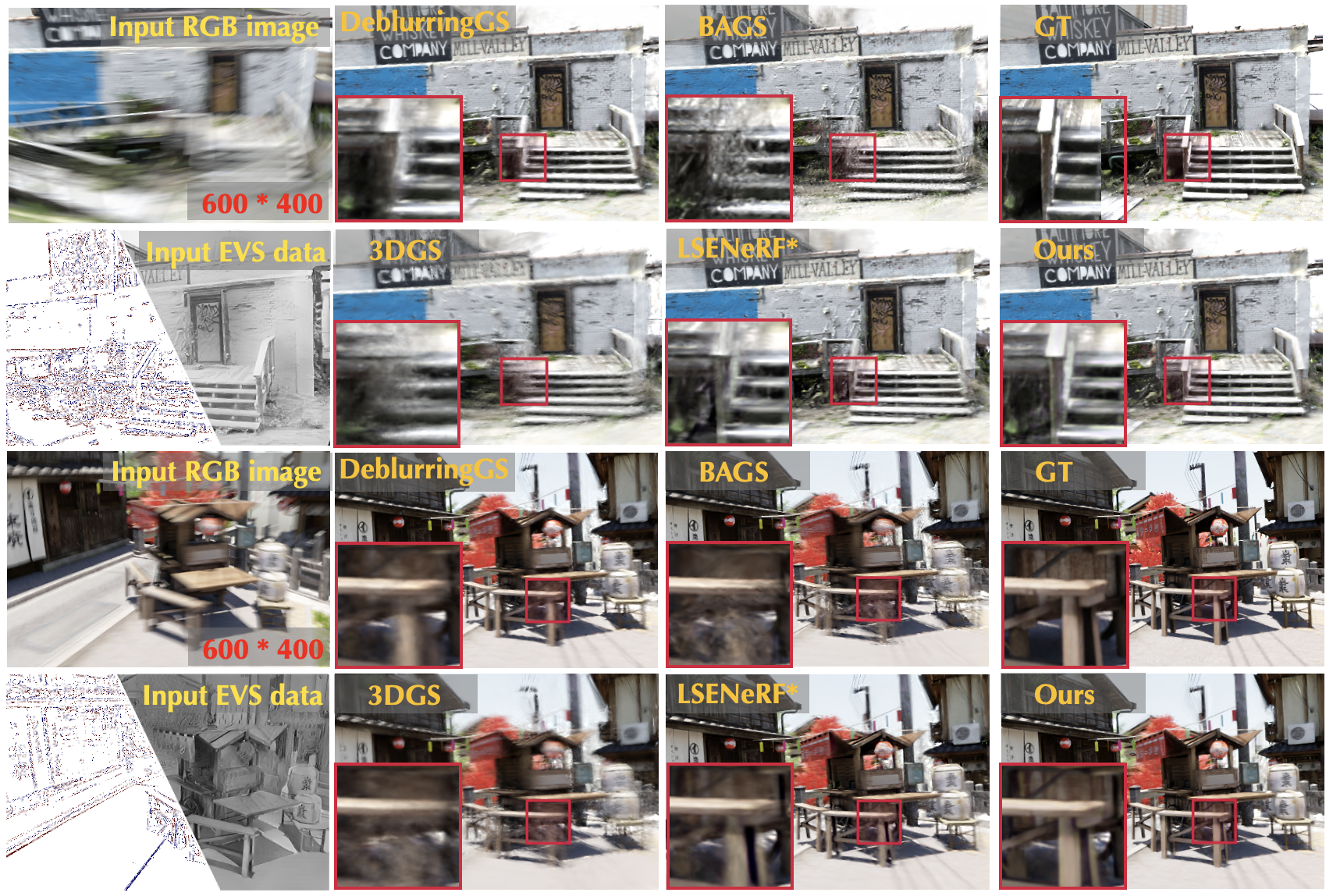}
   \caption{Qualitative comparison on synthetic data, \textit{factory} (top) and \textit{trollet} (bottom). Our method recovers sharp details, such as the stair in the first example, as well as accurate colors, outperforming other event-based and RGB-only methods.}
   \label{fig:synthetic_results}
\end{figure}

\paragraph{Evaluation results.} \cref{tab:synthetic_results} and \cref{fig:synthetic_results} present the quantitative and qualitative evaluations on the synthetic Ev-DeblurBlender dataset. As evidenced by the poor performance of original 3DGS, reconstruction quality is significantly degraded by motion blur. Baselines relying solely on RGB images (\eg, DeblurringGS) offer limited mitigation for this ill-posed problem. In contrast, incorporating motion cues from event cameras yields higher-quality reconstructions. Our method achieves the best performance across all metrics.

\begin{table}[htbp]
\centering
\caption{Quantitative results on the Ev-DeblurBlender dataset. We color code the best \textcolor{blue!70}{\textbf{PSNR}}, \textcolor{yellow!70}{\textbf{SSIM}}, and \textcolor{green!70}{\textbf{LPIPS}} performances.}
\label{tab:synthetic_results}
\resizebox{\textwidth}{!}{
\begin{tabular}{l|ccc ccc ccc ccc ccc}
\toprule
Scene &
\multicolumn{3}{c}{Original 3D GS} &
\multicolumn{3}{c}{BAGS} &
\multicolumn{3}{c}{DeblurringGS} &
\multicolumn{3}{c}{LSENeRF*} &
\multicolumn{3}{c}{Ours} \\
\cmidrule(lr){2-4} \cmidrule(lr){5-7} \cmidrule(lr){8-10} \cmidrule(lr){11-13} \cmidrule(lr){14-16}
& PSNR & SSIM & LPIPS & PSNR & SSIM & LPIPS & PSNR & SSIM & LPIPS & PSNR & SSIM & LPIPS & PSNR & SSIM & LPIPS \\
\midrule
Factory & 18.01 & 0.514 & 0.446 & 19.12 & 0.634 & 0.312 & 21.26 & 0.620 & 0.232 & 20.47 & 0.707 & 0.219 & \cellcolor{blue!20}\textbf{23.18} & \cellcolor{yellow!20}\textbf{0.830} & \cellcolor{green!20}\textbf{0.165} \\
Pool & 17.24 & 0.439 & 0.599 & 23.84 & 0.640 & 0.339 & 15.20 & 0.015 & 0.739 & 24.22 & 0.639 & 0.280 & \cellcolor{blue!20}\textbf{25.16} & \cellcolor{yellow!20}\textbf{0.696} & \cellcolor{green!20}\textbf{0.252} \\
Tanabata & 17.51 & 0.517 & 0.487 & 18.31 & 0.596 & 0.381 & 19.58 & 0.556 & 0.256 & 19.60 & 0.659 & 0.257 & \cellcolor{blue!20}\textbf{20.11} & \cellcolor{yellow!20}\textbf{0.727} & \cellcolor{green!20}\textbf{0.221} \\
Trolley & 18.52 & 0.609 & 0.413 & 20.25 & 0.722 & 0.287 & 21.08 & 0.659 & 0.204 & 20.75 & 0.753 & 0.184 & \cellcolor{blue!20}\textbf{23.49} & \cellcolor{yellow!20}\textbf{0.854} & \cellcolor{green!20}\textbf{0.135} \\
\midrule
Average & 17.82 & 0.520 & 0.486 & 20.38 & 0.648 & 0.330 & 19.28 & 0.463 & 0.358 & 21.26 & 0.689 & 0.235 & \cellcolor{blue!20}\textbf{22.99} & \cellcolor{yellow!20}\textbf{0.777} & \cellcolor{green!20}\textbf{0.193} \\
\bottomrule
\end{tabular}
}
\end{table}
\begin{table}[htbp]
\centering
\caption{Quantitative results on our AsyncEv-Deblur dataset. We color code the best \textcolor{blue!70}{\textbf{PSNR}}, \textcolor{yellow!70}{\textbf{SSIM}}, and \textcolor{green!70}{\textbf{LPIPS}} performances.}
\label{tab:our_results}
\resizebox{\textwidth}{!}{
\begin{tabular}{l|ccc ccc ccc ccc ccc}
\toprule
Scene &
\multicolumn{3}{c}{Original 3D GS} &
\multicolumn{3}{c}{BAGS} &
\multicolumn{3}{c}{DeblurringGS} &
\multicolumn{3}{c}{LSENeRF*} &
\multicolumn{3}{c}{Ours} \\
\cmidrule(lr){2-4} \cmidrule(lr){5-7} \cmidrule(lr){8-10} \cmidrule(lr){11-13} \cmidrule(lr){14-16}
& PSNR & SSIM & LPIPS & PSNR & SSIM & LPIPS & PSNR & SSIM & LPIPS & PSNR & SSIM & LPIPS & PSNR & SSIM & LPIPS \\
\midrule
Patio  & 22.59 & 0.744 & 0.382 & 23.41 & 0.779 & 0.302 & 23.07 & 0.628 & 0.289 & 23.47 & 0.779 & 0.273 & \cellcolor{blue!20}\textbf{24.45} & \cellcolor{yellow!20}\textbf{0.835} & \cellcolor{green!20}\textbf{0.223} \\
Bin    & 22.26 & 0.804 & 0.412 & 25.01 & 0.819 & 0.316 & 23.69 & 0.633 & 0.281 & \cellcolor{blue!20}\textbf{25.88} & 0.827 & \cellcolor{green!20}\textbf{0.258} & 25.67 & \cellcolor{yellow!20}\textbf{0.829} & 0.265 \\
Lounge & 22.20 & 0.778 & 0.434 & 24.14 & 0.834 & 0.347 & \cellcolor{blue!20}\textbf{26.34} & 0.764 & \cellcolor{green!20}\textbf{0.182} & 25.47 & 0.852 & 0.256 & 25.91 & \cellcolor{yellow!20}\textbf{0.881} & 0.199 \\
Bench  & 22.95 & 0.815 & 0.373 & 23.03 & 0.824 & 0.326 & 24.19 & 0.684 & 0.192 & 24.48 & 0.846 & 0.217 & \cellcolor{blue!20}\textbf{27.77} & \cellcolor{yellow!20}\textbf{0.896} & \cellcolor{green!20}\textbf{0.176} \\
Stair  & 24.89 & 0.833 & 0.376 & 26.51 & 0.863 & 0.294 & 27.95 & 0.792 & 0.173 & 26.25 & 0.868 & 0.202 & \cellcolor{blue!20}\textbf{28.43} & \cellcolor{yellow!20}\textbf{0.900} & \cellcolor{green!20}\textbf{0.169} \\
Bus    & 20.53 & 0.737 & 0.466 & 22.06 & 0.762 & 0.403 & 23.43 & 0.615 & 0.255 & 22.14 & 0.760 & 0.295 & \cellcolor{blue!20}\textbf{23.94} & \cellcolor{yellow!20}\textbf{0.808} & \cellcolor{green!20}\textbf{0.251} \\
Wall   & 23.42 & 0.707 & 0.466 & 24.12 & 0.704 & 0.242 & 18.01 & 0.312 & 0.513 & 25.14 & 0.749 & 0.225 & \cellcolor{blue!20}\textbf{25.82} & \cellcolor{yellow!20}\textbf{0.785} & \cellcolor{green!20}\textbf{0.224} \\
\midrule
Average & 22.69 & 0.774 & 0.416 & 24.04 & 0.798 & 0.319 & 23.81 & 0.633 & 0.269 & 24.69 & 0.812 & 0.247 & \cellcolor{blue!20}\textbf{26.00} & \cellcolor{yellow!20}\textbf{0.847} & \cellcolor{green!20}\textbf{0.215} \\
\bottomrule
\end{tabular}
}
\end{table}
\begin{figure}[tb]
  \centering
   \includegraphics[width=1.0\linewidth]{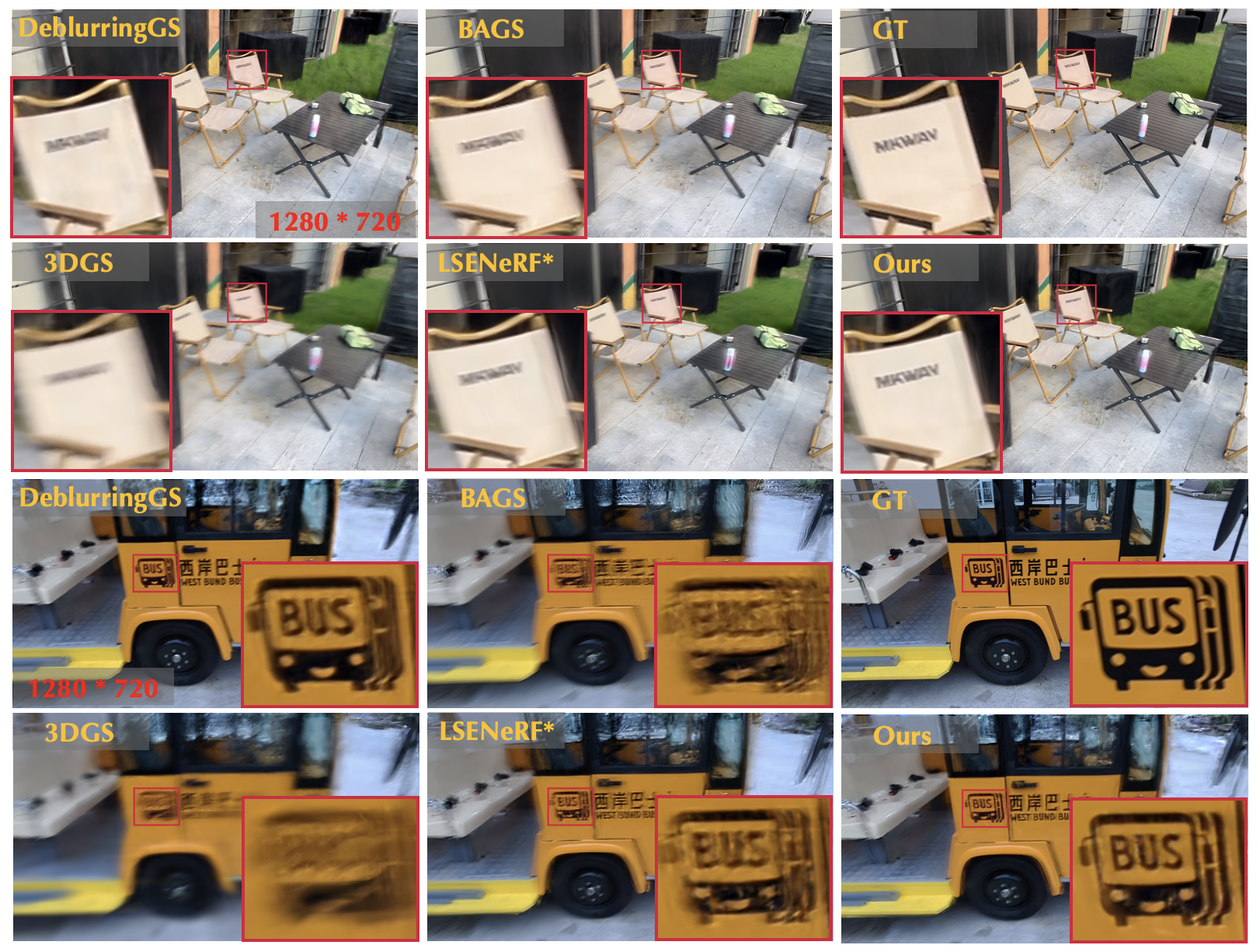}
   \caption{Qualitative comparison on real-world camera motion blur, \textit{Patio} (top) and \textit{Bus} (bottom). Our method recovers high frequency details, such as the text in \textit{Patio} and the logo in  \textit{Bus}.}
   \label{fig:real_world_results}
\end{figure}
\cref{tab:our_results} and \cref{fig:real_world_results} further validate our method on the real-world AsyncEv-Deblur dataset. The performance trends are consistent: baselines relying solely on RGB images suffer from blur artifacts, while event-based supervision substantially improves reconstruction quality. A key benefit is that event-based supervision (shared by LSENeRF~\cite{tang2025lse} and our method) effectively mitigates the edge artifacts introduced by the deblurring module. However, our proposed event structure loss provides more direct supervision than classical event losses. It avoids the ill-posedness associated with calculations between adjacent frames and enables a more comprehensive utilization of the event signals, leading to superior reconstruction quality.

\begin{figure}[!htbp]
  \centering
  \includegraphics[width=\linewidth]{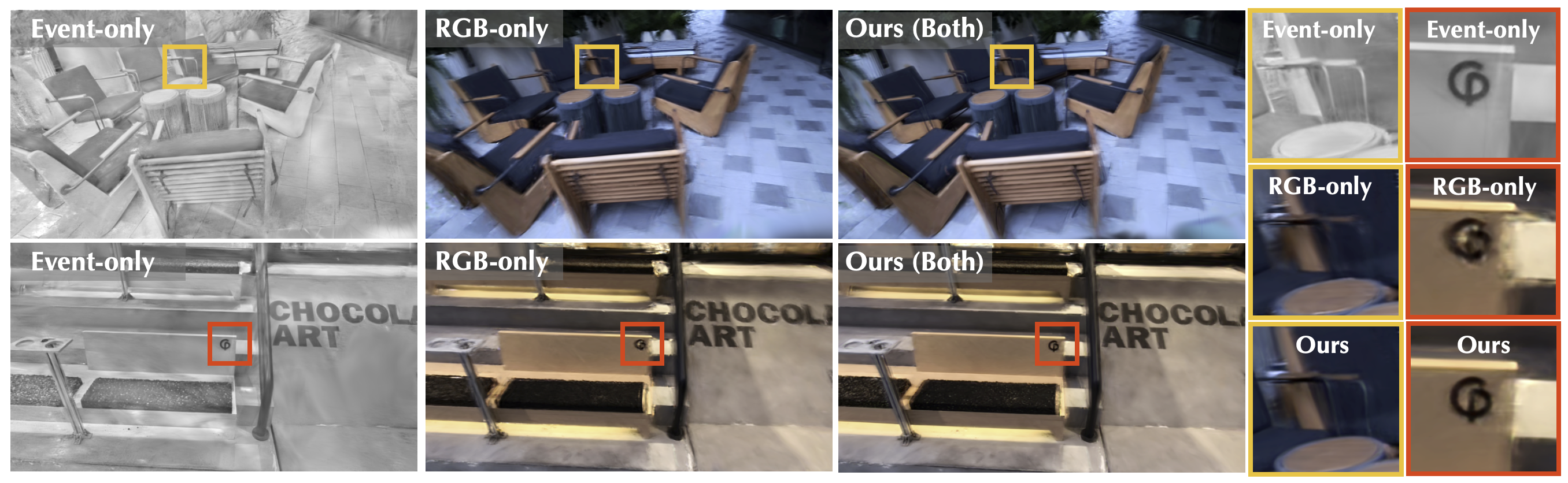}
  \caption{Qualitative ablation on input modalities. \textbf{Event-only} reconstruction captures fine-grained structural details but lacks color information. \textbf{RGB-only} reconstruction preserves color fidelity but suffers from severe blur artifacts. \textbf{Ours (Both)} combines both modalities, achieving sharp details with faithful color reproduction. Zoom-in patches (right) highlight the complementary strengths of each modality.}
  \label{fig:components_ablation}
\end{figure}

\begin{figure}[!t]
  \centering
  \begin{subfigure}[t]{0.48\textwidth}
    \centering
    \includegraphics[width=\linewidth]{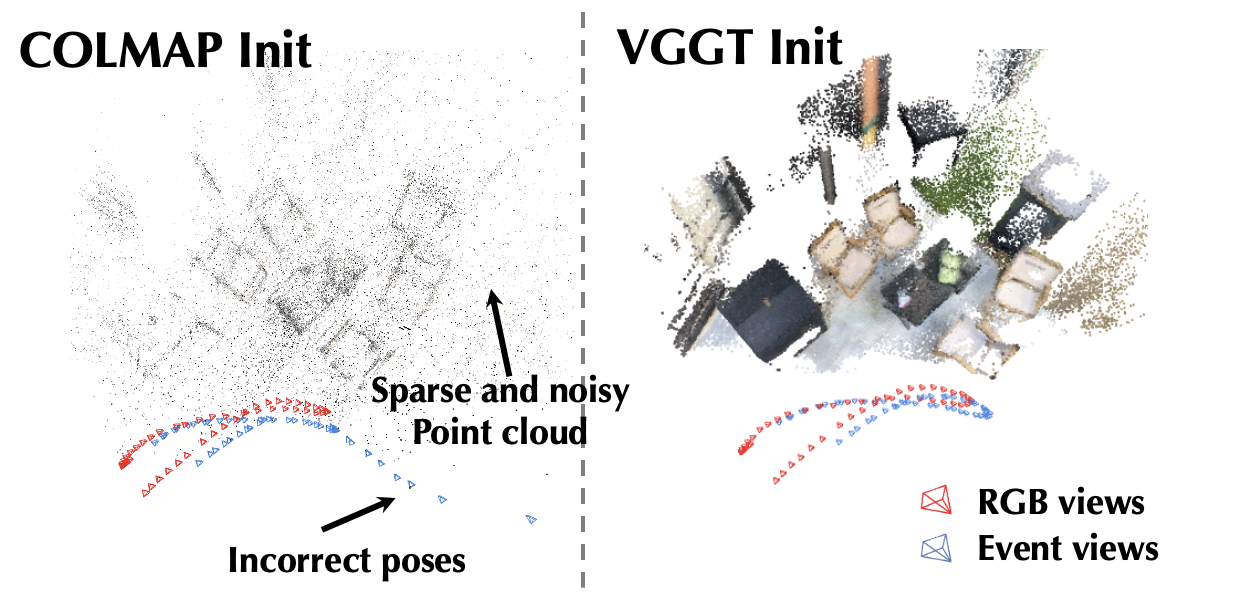}
    \caption{Comparison of initialization methods. COLMAP (left) fails under motion blur, producing sparse, noisy points. Our VGGT-based method (right) robustly estimates accurate poses and generates a dense point cloud.}
    \label{fig:dense_init}
  \end{subfigure}
  \hfill
  \begin{subfigure}[t]{0.48\textwidth}
    \centering
    \includegraphics[width=\linewidth]{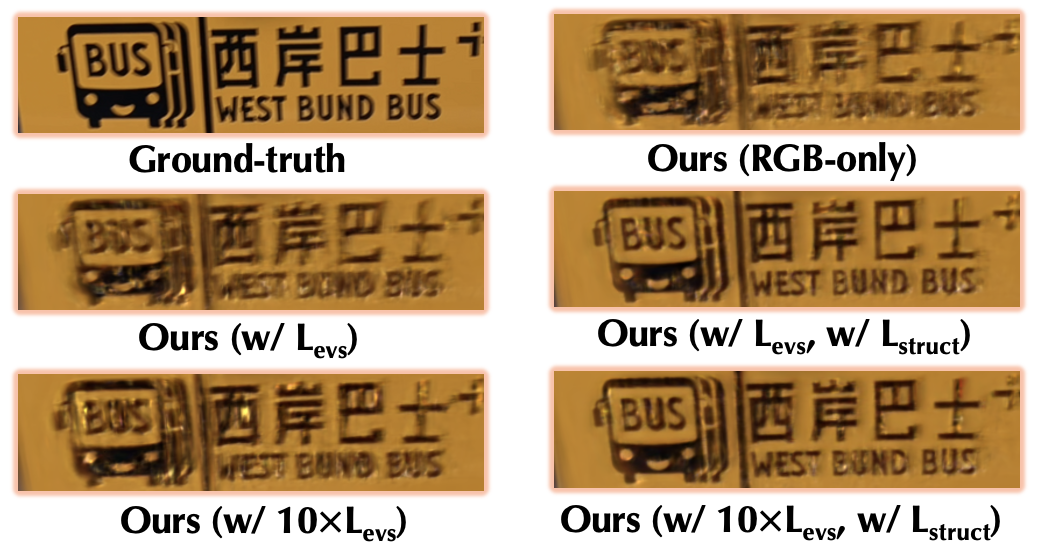}
    \caption{Qualitative ablation of our event loss. RGB-only or traditional $\mathcal{L}_{\text{evs}}$ fails to recover sharp details; increasing its weight ($10\times$) does not help. Only our $\mathcal{L}_{\text{struct}}$ recovers fine, sharp text.}
    \label{fig:loss_ablation}
  \end{subfigure}
  \caption{Qualitative analysis of key components.
    \textbf{(a)} Our VGGT-based initialization is robust to severe motion blur and cross-domain inputs, providing dense, high-quality 3DGS initialization compared to COLMAP.
    \textbf{(b)} Our proposed event structure loss $\mathcal{L}_{\text{struct}}$ successfully incorporates high-frequency event details, outperforming the classical event loss.}
  \label{fig:init-and-loss-ablation}
\end{figure}

\paragraph{Complementary strengths of RGB and Event modalities.}
A core design principle of our method is to exploit the complementary strengths inherent to each sensor modality. As illustrated in \cref{fig:components_ablation}, reconstructing from RGB images alone preserves color fidelity but yields severely blurred results, since the RGB camera inevitably suffers from motion blur during rapid handheld capture. Conversely, the event camera is inherently blur-free and thus captures sharp, high-frequency textures, yet its reconstructions lack color information entirely. Our method effectively fuses both modalities: the event structure loss transfers fine-grained details from the event stream, while the color consistency regularization preserves the rich chromatic information from the RGB frames. The result is a high-quality 3D reconstruction that simultaneously achieves color-accurate appearance and sharp structural detail.
\paragraph{VGGT Initialization.} Obtaining robust camera poses from our cross-domain data, consisting of blurred RGB and event-reconstructed grayscale images, is a key bottleneck for traditional SfM pipelines like COLMAP. As shown in \cref{tab:pose_suceess_rates}, which details the percentage of images each method successfully registered, COLMAP exhibits low registration rates on the blurred RGB frames, failing to provide a complete set of camera poses. In contrast, our VGGT-based initialization demonstrates strong robustness, successfully registering all RGB and event-reconstructed frames and demonstrating its generalization to these challenging, heterogeneous sources.
VGGT's advantage also extends to the quality of the initial point cloud. As illustrated in \cref{fig:dense_init}, the COLMAP initialization leads to incorrect camera pose estimations and a sparse and noisy point cloud. This provides a poor initialization that is insufficient to guide the downstream optimization. In contrast, VGGT directly produces a dense and spatially coherent point cloud even from the blurred inputs. This dense initialization is crucial for preserving geometric continuity and constraining the early 3DGS optimization. Although these initial poses are sufficiently accurate to bootstrap the reconstruction, they are not perfect, so we jointly refine all poses during the 3DGS optimization stage.
\begin{table}[htbp]
\caption{Ablation studies and comparisons.
\textbf{(a)} Pose estimation success rates (\%) for COLMAP and VGGT.
\textbf{(b)} Loss component ablation on the synthetic dataset (Average).
\textbf{(c)} Comparison with 2D deblurring + 3DGS pipelines (Average).
We color code the best \textcolor{blue!70}{\textbf{PSNR$\uparrow$}}, \textcolor{yellow!70}{\textbf{SSIM$\uparrow$}}, and \textcolor{green!70}{\textbf{LPIPS$\downarrow$}}.}
\label{tab:ablation_combined}
\label{tab:pose_suceess_rates}
\label{tab:ablation_loss_quantitative}
\label{tab:2d_deblur_comparison}
\small
\centering
\begin{minipage}[c]{0.48\textwidth}
\centering
\textbf{(a) Pose Estimation Success Rates}
\vspace{2pt}

\resizebox{\textwidth}{!}{%
\begin{tabular}{llccccccc}
\toprule
Method & Cam & Patio & Bin & Lounge & Bench & Stair & Bus & Wall \\
\midrule
\multirow{2}{*}{COLMAP} & RGB   & 96  & 84  & 44  & 36  & 64  & 48  & 92 \\
                        & Event & 85  & 100 & 100 & 100 & 100 & 95  & 100 \\
\midrule
\multirow{2}{*}{VGGT}   & RGB   & 100 & 100 & 100 & 100 & 100 & 100 & 100 \\
                        & Event & 100 & 100 & 100 & 100 & 100 & 100 & 100 \\
\bottomrule
\end{tabular}%
}

\vspace{8pt}
\textbf{(c) 2D Deblurring + 3DGS}
\vspace{2pt}

\resizebox{\textwidth}{!}{%
\begin{tabular}{l ccc}
\toprule
Method & PSNR$\uparrow$ & SSIM$\uparrow$ & LPIPS$\downarrow$ \\
\midrule
NAFNet~\cite{chen2022simple} + 3DGS & 22.72 & 0.778 & 0.399 \\
Restormer~\cite{zamir2022restormer} + 3DGS & 22.79 & 0.778 & 0.403 \\
ShiftNet~\cite{li2023shiftnet} + 3DGS & 21.84 & 0.750 & 0.415 \\
\midrule
Ours & \cellcolor{blue!20}\textbf{26.00} & \cellcolor{yellow!20}\textbf{0.847} & \cellcolor{green!20}\textbf{0.215} \\
\bottomrule
\end{tabular}%
}
\end{minipage}
\hfill
\begin{minipage}[c]{0.50\textwidth}
\centering
\textbf{(b) Loss Component Ablation}
\vspace{2pt}

\resizebox{\textwidth}{!}{%
\begin{tabular}{l ccc}
\toprule
Configuration & PSNR$\uparrow$ & SSIM$\uparrow$ & LPIPS$\downarrow$ \\
\midrule
w/o pose optim. & 21.26 & 0.691 & 0.217 \\
\midrule
w/o $\mathcal{L}_{\text{reg-r}}$ & 22.33 & 0.773 & 0.199 \\
w/o $\mathcal{L}_{\text{struct}}$ & 22.26 & 0.751 & 0.219 \\
w/o $\mathcal{L}_{\text{reg-e}}$ & 22.78 & 0.777 & 0.198 \\
\midrule
LSENeRF* \cite{tang2025lse} & 21.26 & 0.689 & 0.235 \\
Ours (Full) & \cellcolor{blue!20}\textbf{22.99} & \cellcolor{yellow!20}\textbf{0.777} & \cellcolor{green!20}\textbf{0.193} \\
\bottomrule
\end{tabular}%
}
\end{minipage}
\end{table}
\paragraph{Impact of loss functions.}
\label{sec:ablation_loss}
We conduct a comprehensive ablation study to validate the design of our multi-objective loss function, with quantitative results in \cref{tab:ablation_combined} and qualitative insights in \cref{fig:loss_ablation}. Our analysis confirms that our full model significantly outperforms all ablated variants. We validate our two primary algorithmic contributions: the two-part color consistency regularizer ($\mathcal{L}_{\text{reg-r}}, \mathcal{L}_{\text{reg-e}}$) and the event structure loss ($\mathcal{L}_{\text{struct}}$). Quantitatively, removing either of the color regularization components degrades performance, validating our claim in \cref{sec:color_reg} that they are crucial for stabilizing the deblurring module and maintaining global color consistency. Disabling our $\mathcal{L}_{\text{struct}}$ results in an even more significant performance drop, confirming its vital role in integrating high-frequency event details.
Qualitatively, \cref{fig:loss_ablation} reinforces our motivation for $\mathcal{L}_{\text{struct}}$ (from \cref{sec:event_loss}). Reconstructing with only RGB data fails to recover sharp details from the motion blur. Adding the traditional event loss provides some structural guidance, but the result remains ill-defined.
In contrast, our proposed $\mathcal{L}_{\text{struct}}$ successfully harnesses the high-frequency event data to restore sharp, fine-grained text, proving its superiority over traditional event-based supervision.
We also conduct ablations on the pose optimization, as it is important for the reconstruction. The results confirm that both components indeed provide substantial improvements to the final reconstruction quality.

\begin{figure}[!htbp]
  \centering
  \includegraphics[width=0.75\linewidth]{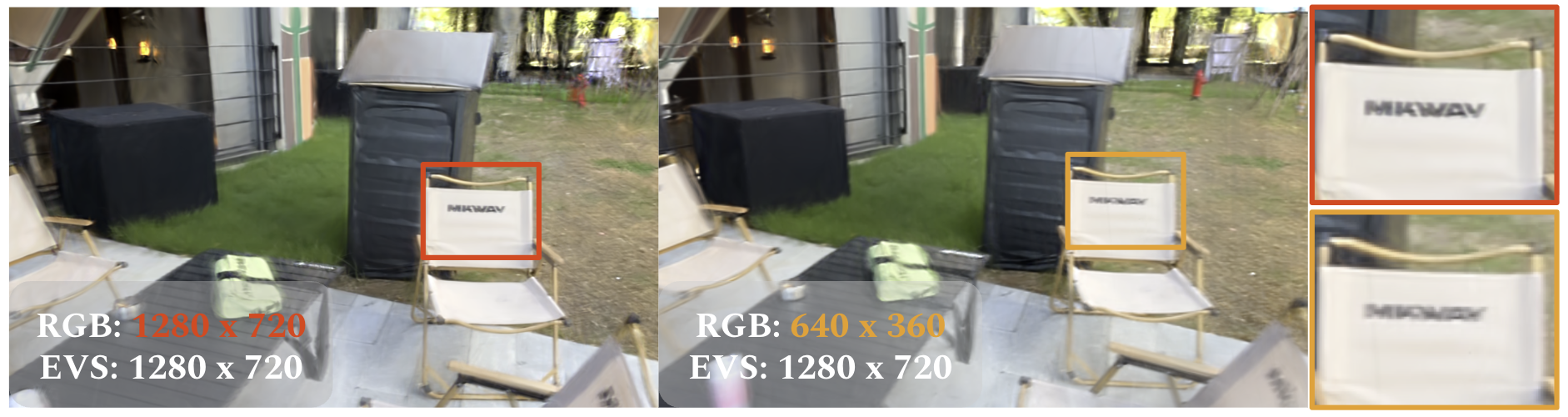}
  \caption{Reconstruction with mismatched resolutions.}
  \label{fig:res_mismatch}
\end{figure}

\paragraph{Comparison with 2D deblurring pipelines.}
An alternative to our end-to-end approach is to first apply a 2D deblurring network per frame and then reconstruct with standard 3DGS. As shown in \cref{tab:2d_deblur_comparison}, this pipeline consistently underperforms our method by a large margin, despite using state-of-the-art deblurring models~\cite{chen2022simple,zamir2022restormer,li2023shiftnet}. The gap arises because (1) 2D deblurring cannot fully recover sharp details under severe motion blur, and (2) per-frame processing provides no multi-view consistency guarantee, leading to degraded 3D reconstruction.
\paragraph{Resolution flexibility.}
Our asynchronous dual-camera setup does not require the two sensors to share the same resolution. To verify this, we halve the RGB resolution to $640\times360$ while keeping the event camera at $1280\times720$. As shown in \cref{fig:res_mismatch}, the pipeline still produces reasonable reconstructions, confirming its flexibility. Nevertheless, the quality degradation compared to the full-resolution setting demonstrates the advantage of our high-resolution system in achieving superior reconstruction quality.
\section{Conclusions}
\label{sec:conclusion}
We introduce a novel, flexible, high-resolution asynchronous RGB-Event dual-camera system that effectively leverages blurry RGB images and high-frame-rate event signals for high-quality 3D reconstruction. Our approach addresses the critical initialization bottleneck—where standard SfM (\eg, COLMAP) fails due to motion blur—by leveraging VGGT for robust cross-domain pose estimation. To optimize the 3DGS representation, we augment traditional event-based losses with a novel event structure loss to robustly harness high-frequency motion details. Furthermore, we introduce a crucial two-part consistency regularizer to prevent deblurring artifacts and distill color to event-only views. This system design facilitates more efficient scene acquisition. Extensive evaluations on both synthetic and real-world datasets demonstrate that our method achieves state-of-the-art reconstruction quality, significantly outperforming existing baselines.

% ...

\bibliographystyle{splncs04}
\bibliography{main}

\clearpage

\renewcommand{\topfraction}{0.9}
\renewcommand{\bottomfraction}{0.9}
\renewcommand{\textfraction}{0.1}
\renewcommand{\floatpagefraction}{0.8}
\setcounter{topnumber}{4}
\setcounter{bottomnumber}{4}
\setcounter{totalnumber}{8}

\setcounter{section}{0}
\renewcommand{\thesection}{\Alph{section}}
\setcounter{figure}{0}
\renewcommand{\thefigure}{S\arabic{figure}}

\setcounter{table}{0}
\renewcommand{\thetable}{S\arabic{table}}

\setcounter{equation}{0}
\renewcommand{\theequation}{S\arabic{equation}}

\section{Working principle of Event camera}
Unlike conventional cameras that capture full frames at a fixed rate, an event camera, also known as a Dynamic Vision Sensor (DVS), is a bio-inspired visual sensor that operates asynchronously. It independently monitors the change in logarithmic intensity for each pixel $\mathbf{x}=(x, y)$.

When the change in log intensity $\mathcal{I}(\mathbf{x}, t)$ at time $t$ exceeds a predefined contrast threshold $\Theta$ compared to the value at the last event $t-\Delta t$ for that pixel, it asynchronously triggers an event $e = (\mathbf{x}, t, p)$. The polarity $p \in \{+1, -1\}$ indicates the direction of the brightness change (increase or decrease). This triggering mechanism is shown in Equation (1):
$$
p =
\begin{cases}
+1, & \text{if } \log(\mathcal{I}(\mathbf{x}, t)) - \log(\mathcal{I}(\mathbf{x}, t-\Delta t)) > \Theta \\
-1, & \text{if } \log(\mathcal{I}(\mathbf{x}, t)) - \log(\mathcal{I}(\mathbf{x}, t-\Delta t)) < -\Theta
\end{cases}
\eqno{(1)}
$$
Therefore, an event camera outputs a spatially sparse but temporally dense (microsecond resolution) stream of events, recording only the dynamic information in the scene, which effectively avoids motion blur and offers a high dynamic range.

\section{More details about the Event Structure Loss}
\subsection{More details about structure extractor}
Our method extracts a high-frequency \textit{structure} component from an input image via local contrast normalization, separating it from low-frequency \textit{color} information.

For an RGB input image, we first convert it to the YUV colorspace. The luminance channel ($Y$) is isolated for structure extraction, while the chrominance channels ($U, V$) are preserved as the color component. If the input is already grayscale, it is processed directly as the luminance channel.

The structure is extracted by standardizing the luminance channel $Y$. We compute local mean $\mu$ and local standard deviation $\sigma$ for each pixel with a Gaussian blur operator ($G_{k, \sigma_X}$) to approximate the local statistics. The structure component $S$ is defined as:
$$
S(\mathbf{x}) = \frac{Y(\mathbf{x}) - \mu(\mathbf{x})}{\sigma(\mathbf{x}) + c}
$$
where $\mathbf{x}$ denotes the pixel coordinates and $c$ is a small constant for stability in low-variance regions.

Finally, this structure component $S$ is normalized to the range $[0, 1]$ to produce the final structure map, $S_{\text{norm}}$. This normalized map is returned along with the color component (if applicable).

\subsection{More details about weight mask}
To generate a weight map that highlights salient and stable features from the event-reconstructed grayscale image $I \in [0, 1]$, we implement a multi-scale gradient analysis pipeline. The goal is to identify structures that are not only strong at a coarse level but also persistent across scales (i.e., not just fine-scale noise).

First, we compute image gradients at two distinct scales. The input image $I$ is convolved with two Gaussian kernels, $G_s$ (with standard deviation $\sigma_{\text{small}}$) and $G_l$ (with $\sigma_{\text{large}}$), to produce a fine-scale version $I_s$ and a coarse-scale version $I_l$. Denoted as:
$
I_s = G_s * I, I_l = G_l * I
$
We then apply Sobel operators ($\nabla_x, \nabla_y$) to both blurred images to obtain their respective gradient magnitudes, $M_s$ and $M_l$:
$$
M_s = \sqrt{(\nabla_x I_s)^2 + (\nabla_y I_s)^2 + \epsilon}
$$
$$
M_l = \sqrt{(\nabla_x I_l)^2 + (\nabla_y I_l)^2 + \epsilon}
$$
where $\epsilon$ is a small constant (e.g., $10^{-12}$) for numerical stability.

To ensure the detected structures are stable and not just fine-scale artifacts, we compute a \textit{cross-scale persistence} score $P$. We first robustly normalize $M_s$ and $M_l$ to range $[0, 1]$ using a normalization function $\mathcal{N}_R(\cdot)$ (which clips outliers and performs min-max scaling), yielding $M'_s$ and $M'_l$. The persistence $P$ is then calculated as ratio of coarse-to-fine magnitude, clamped at $1.0$ and modulated by an exponent $\gamma$:
$$
P = \left( \min\left( \frac{M'_l}{M'_s + \epsilon}, 1.0 \right) \right)^\gamma
$$
This term assigns high scores to structures present at both scales ($M'_l \approx M'_s$) and suppresses features that are strong at fine scale but absent at coarse scale ($M'_l \ll M'_s$).

We further refine the mask by gating out weak structures at the coarse level. A soft gate $G$ is computed using the normalized coarse magnitude $M'_l$:
$
G = \text{sigmoid}(s \cdot (M'_l - \tau)),
$
where $s$ is a sharpness parameter (e.g., $25.0$) and $\tau$ is a dynamic threshold, typically set as the 85th percentile of $M'_l$ (i.e., $\tau = Q_{0.85}(M'_l)$). This gate $G$ effectively binarizes the coarse map, retaining only the most salient features.

The unnormalized weight map $\hat{W}$ is defined as the product of the coarse-scale salience, the cross-scale persistence, and the coarse-scale gate:
$
\hat{W} = M'_l \cdot P \cdot G.
$
This map $\hat{W}$ is then robustly normalized to produce the final weight map $W = \mathcal{N}_R(\hat{W})$.

\begin{figure}[!htbp]
  \centering
  \includegraphics[width=0.9\linewidth]{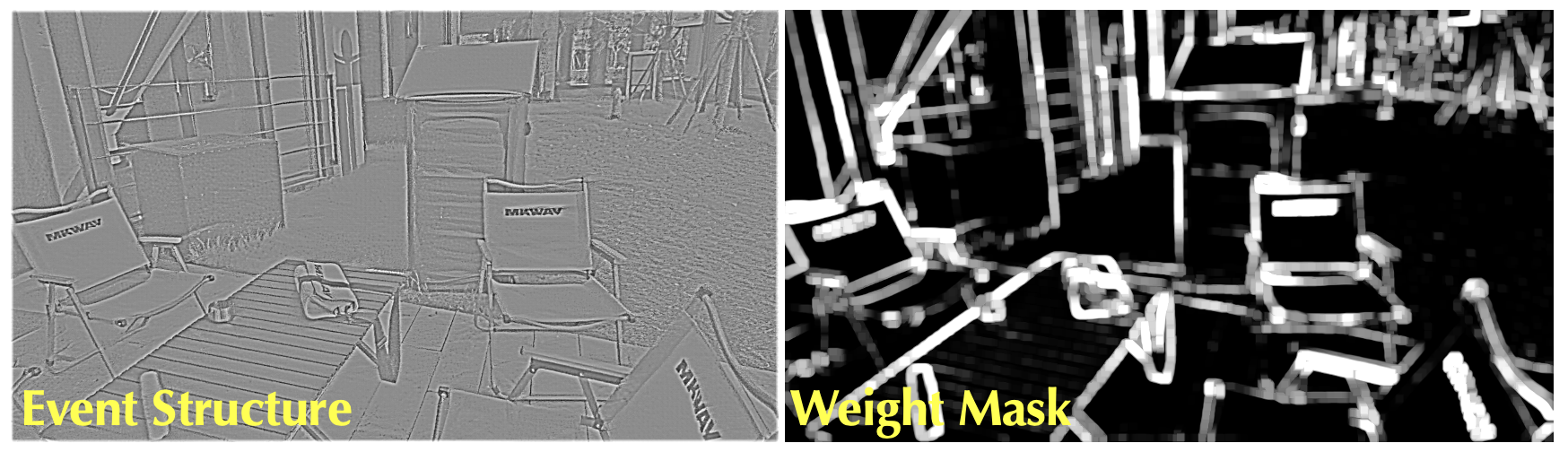}
   \caption{Extracted Event Structure and Weight Mask.}
   \label{fig:supp_weight_mask}
\end{figure}

Optionally, a morphological dilation (implemented as a 2D max-pooling operation with stride 1) with a kernel $k_d$ is applied to $W$ to slightly thicken the resulting structural mask for downstream tasks. We visualize the examples of extracted event structures and created weight mask in \cref{fig:supp_weight_mask}.

\section{More details about Deblurring MLP}
To simulate motion blur within the 3D Gaussian Splatting framework, we utilize a multi-layer perceptron (MLP), adopted from Deblurring 3DGS~\cite{lee2024deblurring}, which predicts a series of deformations for each 3D Gaussian to represent its state at multiple discrete moments during exposure. The final blurred rendering is achieved by averaging these deformed states. The network input is an 85-dimensional feature vector, which concatenates positional embeddings of the 3D position ($L=3$) and 3D view direction ($L=10$), along with the raw 3D scale (3 channels) and 4D rotation quaternion (4 channels). The core MLP (using default parameters \texttt{num\_hidden=3}, \texttt{width=64}) consists of 3 linear layers with ReLU activations, transforming the input into a 64-dimensional feature vector. This feature is then passed to three separate linear output heads to predict the deltas for position (\texttt{self.p}, $3 \times 4 = 12$ channels), scale (\texttt{self.s}, $3 \times 5 = 15$ channels), and rotation (\texttt{self.r}, $4 \times 5 = 20$ channels). The learning rate for the MLP is constantly $0.001$.

\section{Camera Pose Calibration Pipeline Comparison}
\begin{figure}[!htbp]
  \centering
  \includegraphics[width=\linewidth]{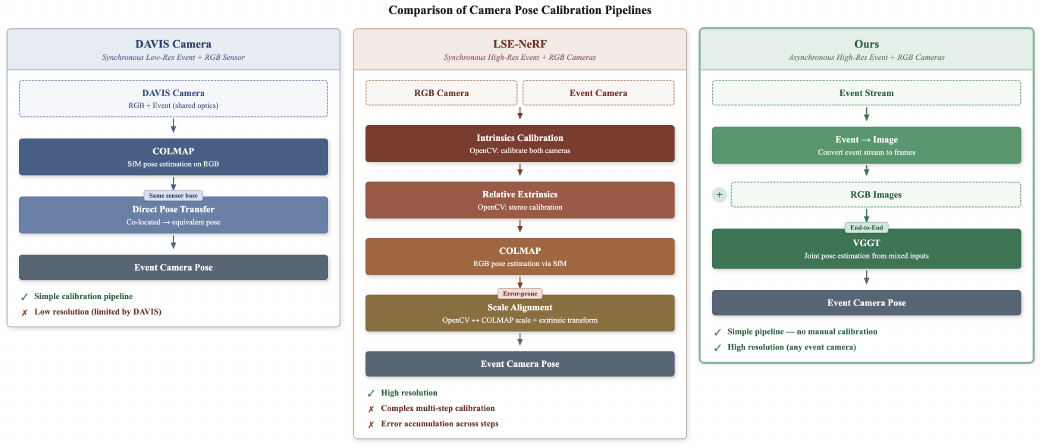}
  \caption{Comparison of camera pose calibration pipelines. Our approach replaces complex multi-step calibration with a single end-to-end VGGT estimation, achieving simplicity, high resolution, and flexibility.}
  \label{fig:supp_pose_pipeline}
\end{figure}

\cref{fig:supp_pose_pipeline} compares the camera pose calibration pipelines of three representative setups. DAVIS-based methods benefit from co-located sensors but are limited to low resolution ($346\times260$). LSE-NeRF achieves high resolution via a synchronized dual-camera rig, but requires a complex multi-step calibration (intrinsics, stereo extrinsics, SfM, and scale alignment), where errors accumulate across stages. In contrast, our method feeds both event-reconstructed frames and RGB images directly into VGGT for end-to-end joint pose estimation, eliminating manual calibration entirely. This simple pipeline supports any event camera at arbitrary resolution, offering both flexibility and robustness.

\section{More Details about AsyncEv-Deblur Dataset.}
We introduce the AsyncEv-Deblur Dataset, which offers two significant advantages over prior work. First, it features a substantially higher resolution. Both our RGB and EVS sensors provide a resolution of $1280 \times 720$, which is significantly higher than the $346 \times 260$ resolution of typical DAVIS cameras. Second, we employ a more flexible asynchronous dual-camera setup that does not require temporal synchronization between the sensors.

Notably, unlike the capture conditions of many previous datasets which focus on low-light, long-exposure scenarios, our data was collected under normal outdoor illumination with standard exposure times. This makes our dataset more representative of real-world use cases and provides a more practical benchmark for evaluation.

\begin{figure}[!htbp]
  \centering
  \includegraphics[width=1.0\linewidth]{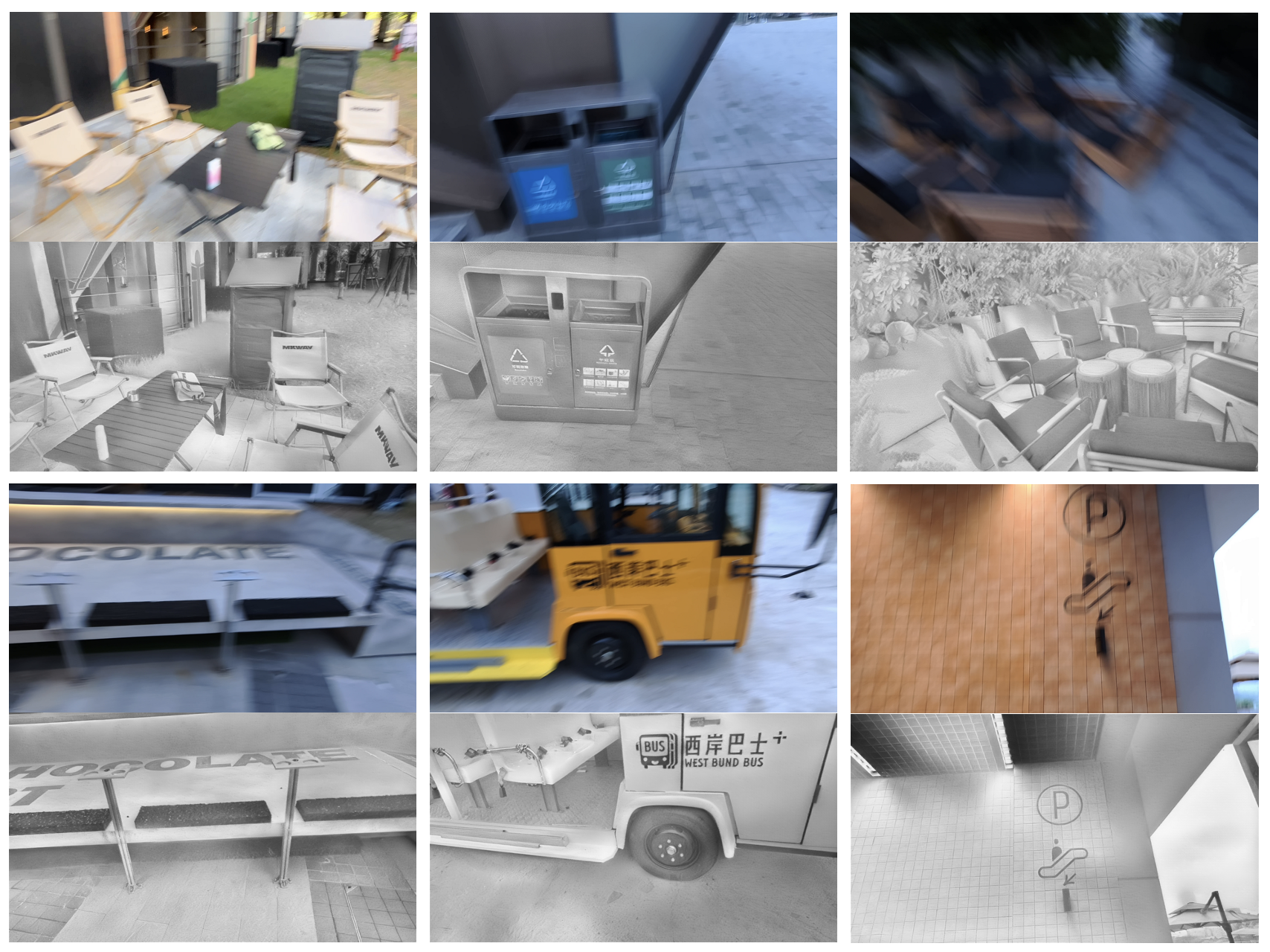}
  \caption{More details about our AsynEv-Deblur datasets. AsynEv-Deblur dataset contains diverse scenes, and captured by our dual-camera system with high capturing speed.}
  \label{fig:supp_datasets}
\end{figure}

The dataset comprises seven distinct scenes, we demonstrate more scenes in \cref{fig:supp_datasets}. For each scene, we provide 25-50 RGB images, which include blurry training views and sharp testing views. Concurrently, we offer a comparable number of event-reconstructed images generated via E2VID. All views were calibrated to obtain camera intrinsics and extrinsics, with initialization performed using VGGT. Furthermore, while our asynchronous setup does not necessitate camera synchronization or timestamp recording, we provide the event camera timestamps and raw event signals for the convenience of future work and comparative analysis.

\section{Event images pre-procession}
\begin{figure}[!htbp]
  \centering
  \includegraphics[width=1.0\linewidth]{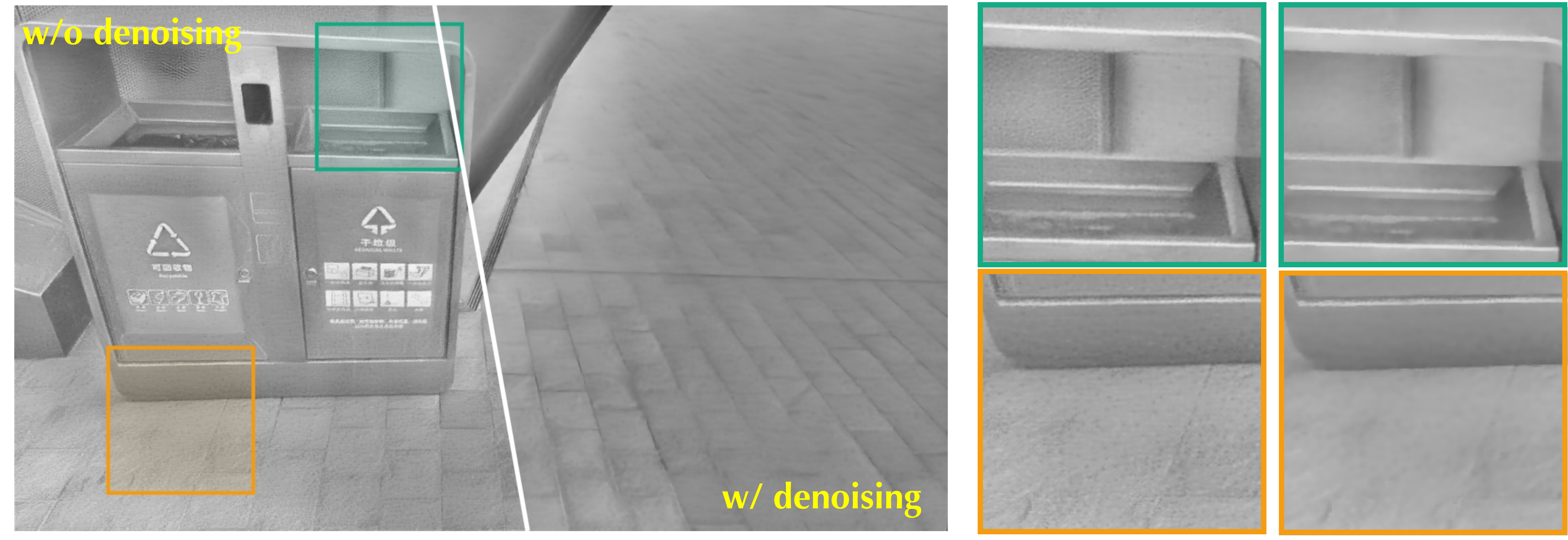}
  \caption{Comparison of bilateral denoising pre-processing.}
  \label{fig:supp_denosing}
\end{figure}

\begin{figure}[!htbp]
  \centering
  \includegraphics[width=0.75\linewidth]{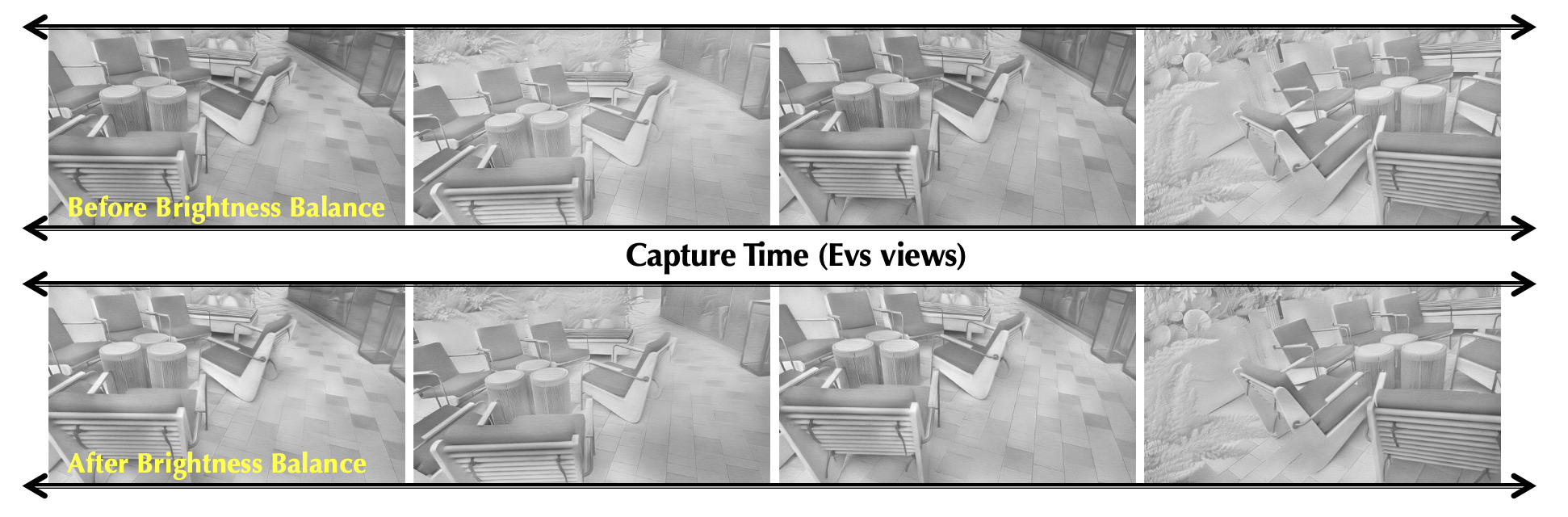}
  \caption{Comparison of brightness balance.}
  \label{fig:supp_brightness}
\end{figure}

The grayscale event images reconstructed via E2VID~\cite{Rebecq19cvpr} exhibit two primary artifacts that degrade downstream performance. First, they suffer from significant noise, as illustrated in~\cref{fig:supp_denosing}, which is a byproduct of unavoidable event noise during real-world capture. Second, the reconstructed sequence displays pronounced inter-frame brightness inconsistencies, particularly in non-edge regions, as shown in~\cref{fig:supp_brightness}.

\begin{figure}[!htbp]
  \centering
  \includegraphics[width=0.75\linewidth]{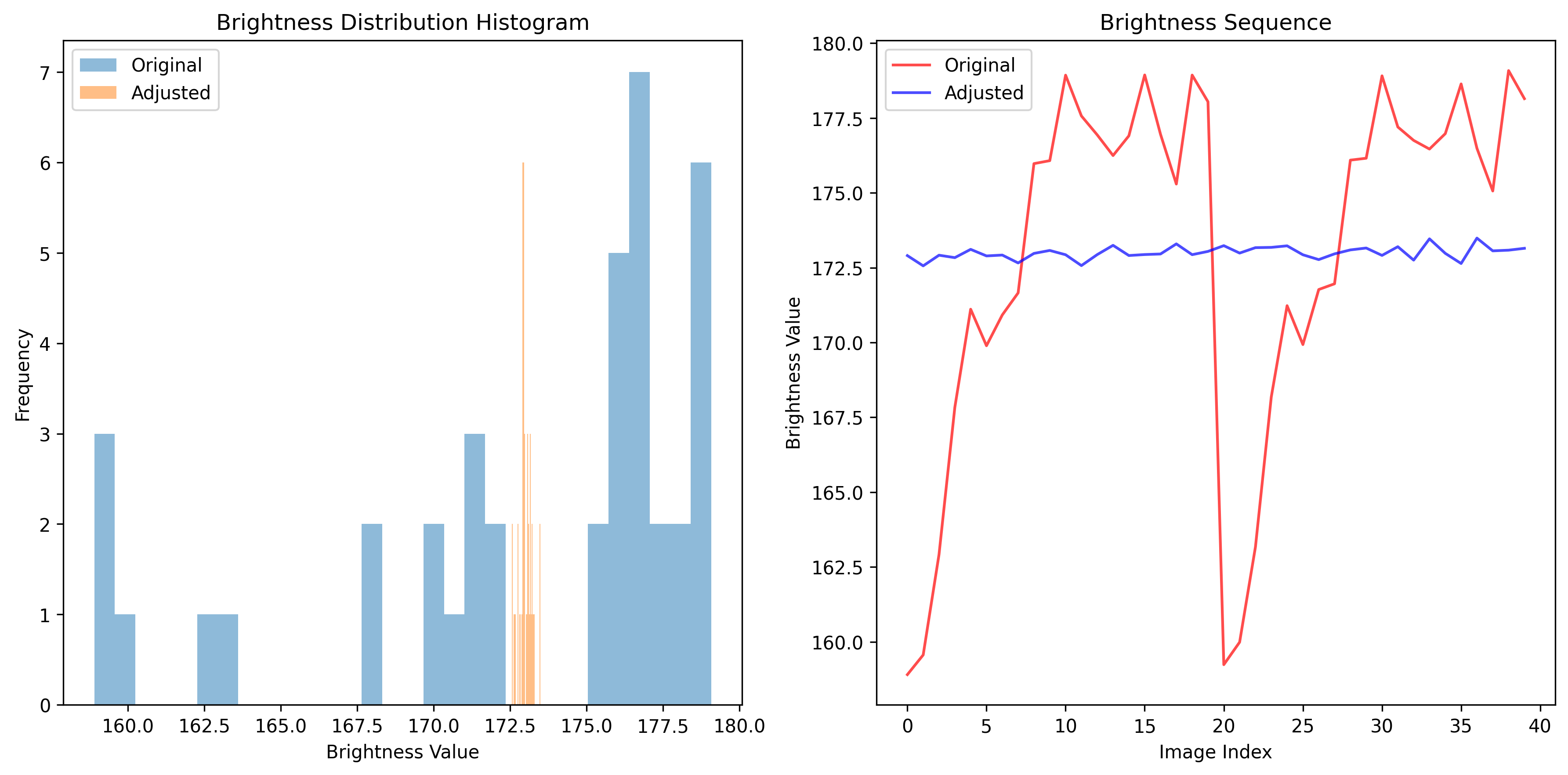}
  \caption{A representative brightness balance pre-processing results.}
  \label{fig:supp_brightness_analysis}
\end{figure}

To mitigate the adverse effects of these issues on subsequent VGGT initialization and Gaussian Splatting reconstruction, we apply a two-stage preprocessing pipeline to all event-reconstructed images. Initially, we apply Bilateral Denoising to all frames. This step markedly reduces the noise level, as shown in~\cref{fig:supp_denosing}. Subsequently, to enhance temporal photometric consistency, we perform a brightness balancing procedure across the image sequence. As demonstrated in~\cref{fig:supp_brightness} and~\cref{fig:supp_brightness_analysis}, this step significantly alleviates inter-frame brightness inconsistency, thereby reducing potential artifacts in the final reconstruction.

\begin{figure}[!htbp]
  \centering
  \includegraphics[width=0.85\linewidth]{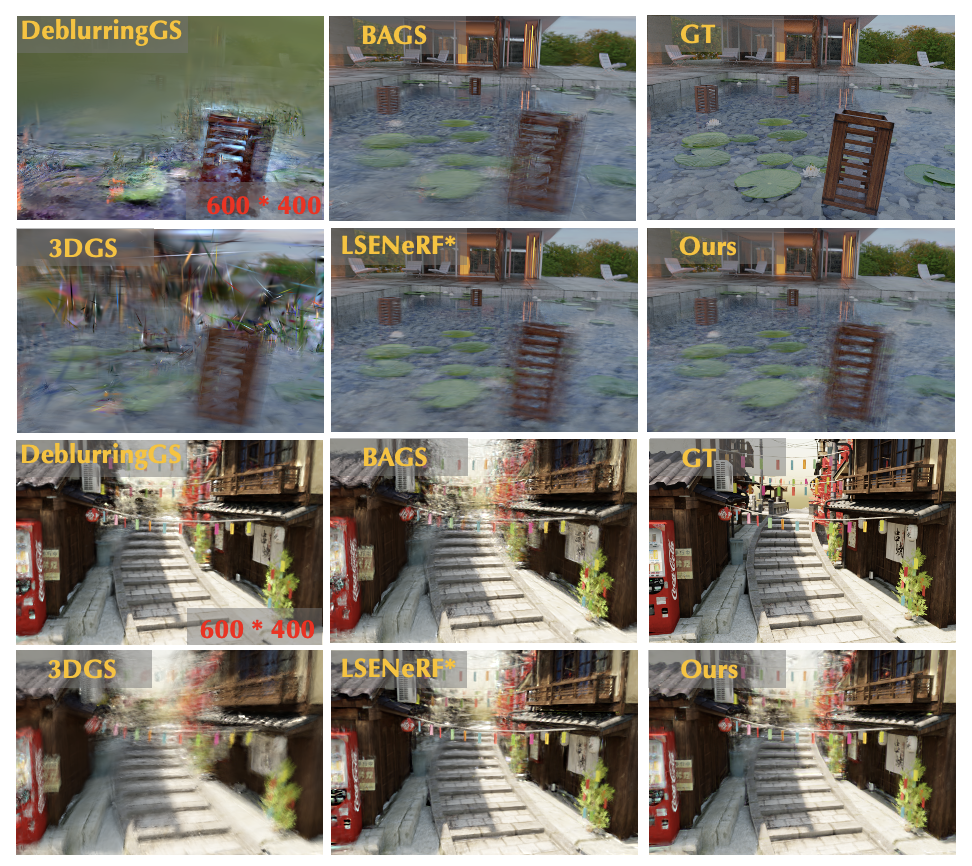}
  \caption{More qualitative results on Ev-Deblu Blender dataset.}
  \label{fig:supp_more_results_synthetic}
\end{figure}

\begin{figure}[!htbp]
  \centering
  \includegraphics[width=0.85\linewidth]{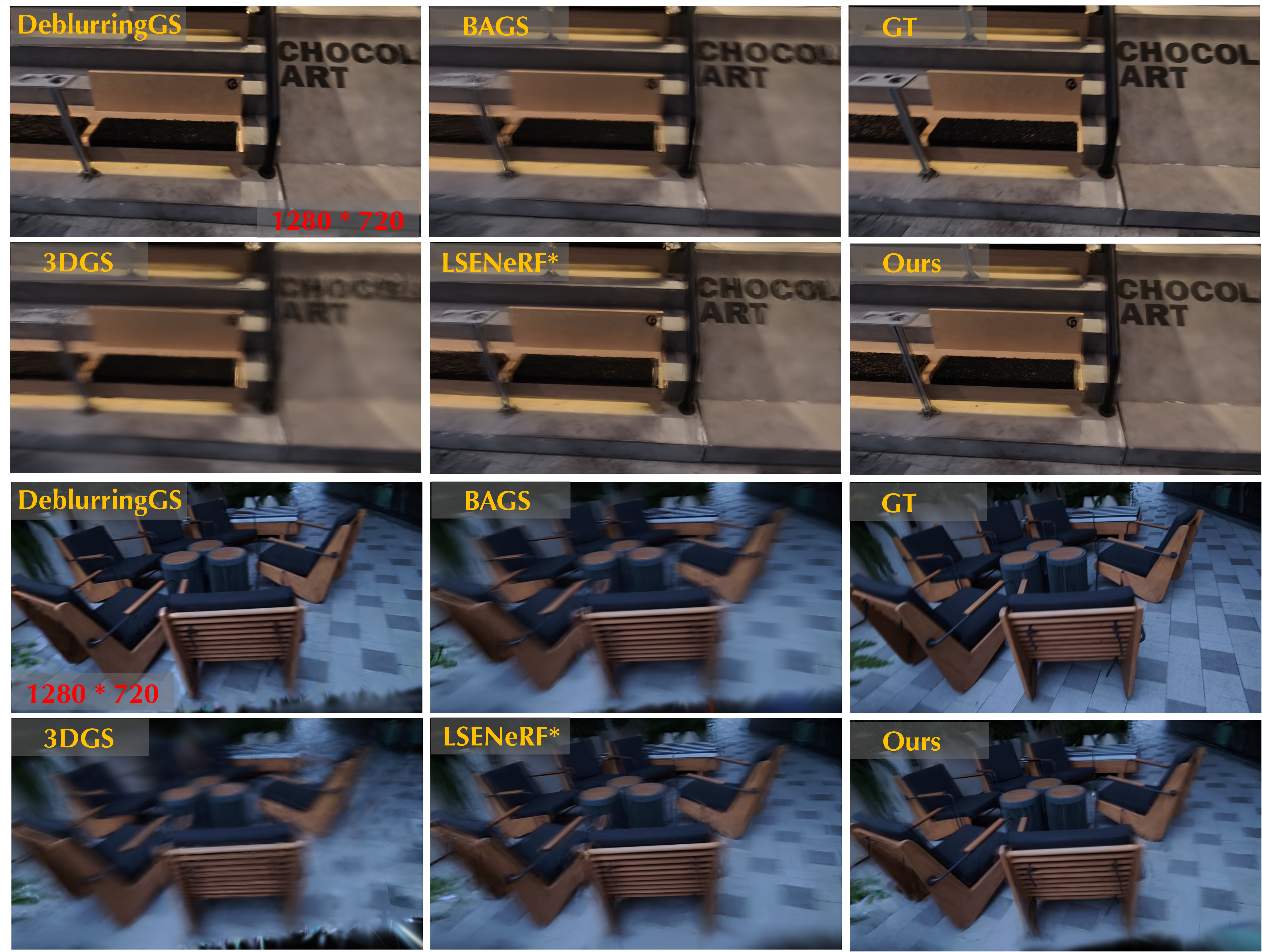}
  \caption{More qualitative results on our AsyncEv-Deblur dataset.}
  \label{fig:supp_more_results}
\end{figure}

\end{document}